\documentclass[10pt,oneside]{article}

\usepackage[a4paper, total={6in, 9in}]{geometry} %

\usepackage{natbib} 
\usepackage{xcolor}
\usepackage{graphicx}
\usepackage{xspace}
\usepackage{eso-pic}
\usepackage[colorlinks,urlcolor=blue,citecolor=blue,bookmarks=false]{hyperref}

\usepackage{amsmath,amssymb,amsfonts,amsthm}
\usepackage{mathtools}
\usepackage{comment}
\usepackage{color}
\usepackage{booktabs}
\usepackage{multirow}

\usepackage{cleveref}
\usepackage[figuresleft]{rotating}

\DeclareMathOperator{\argmin}{\arg\min}

\usepackage{todonotes}

\usepackage{dsfont}

\newcommand*\samethanks[1][\value{footnote}]{\footnotemark[#1]}

\usepackage{xcolor}

\begin{document}

\title{Is Differentiable Architecture Search truly a One-Shot Method?}

\author{
  Jonas Geiping\thanks{Authors contributed equally.}\\
  \texttt{jonas.geiping@uni-siegen.de}
  \and
  Jovita Lukasik\samethanks\\
  \texttt{jovita@uni-mannheim.de}
  \and
  Michael Moeller\thanks{Authors contributed equally.}\\
  \texttt{michael.moeller@uni-siegen.de}
  \and
  Margret Keuper\samethanks\\
  \texttt{margret.keuper@uni-siegen.de}
 \\  
}

\maketitle

\begin{abstract}
Differentiable architecture search (DAS) is a widely researched tool for the discovery of novel architectures, due to its promising results for image classification. The main benefit of DAS is the effectiveness achieved through the weight-sharing one-shot paradigm, which allows efficient architecture search.
In this work, we investigate DAS in a systematic case study of inverse problems, which allows us to analyze these potential benefits in a controlled manner. 
We demonstrate that the success of DAS can be extended from image classification to signal reconstruction, in principle. However, our experiments also expose three fundamental difficulties in the evaluation of DAS-based methods in inverse problems: First, the results show a large variance in all test cases. Second, the final performance is strongly dependent on the hyperparameters of the optimizer. And third, the performance of the weight-sharing architecture used during training does not reflect the final performance of the found architecture well. 
While the results on image reconstruction confirm the potential of the DAS paradigm, they challenge the common understanding of DAS as a one-shot method.
\end{abstract}


\section{Introduction}\label{sec1}

Recent progress in computer vision and related fields
has illustrated the importance of suitable neural architecture designs and training schemes \cite{he_deep_2015}. Ever deeper and more complex networks show promise, and manual network design is less and less able to explore the desired search spaces.
Neural architecture search (NAS) is the task of optimizing the architecture of a neural network automatically without resorting to human selection, scaling to larger search spaces and proposing novel well-performing architectures.
NAS, which is an intrinsically discrete problem, has been successfully addressed using black-box optimization approaches such as reinforcement learning~\cite{Zoph2017,zoph2018learning} or Bayesian optimization~\cite{Kandasamy2018,white2019bananas,Ru2020NeuralAS,Lukasik_2021_svge}. However, these approaches are computationally expensive as they require the training of many candidate networks to cover the search space. 
In contrast, differentiable architecture search (DAS) \cite{liu_darts_2019} proposes a continuous relaxation of the search problem, i.e.~all candidate architectures within a given search space of operations and their connectivity are jointly optimized using shared network parameters while the network also learns to weigh these operations. The final architecture can then simply be deduced by selecting the highest weighted operations.
This is appealing as practically good architectures are proposed within a single optimization run. However, previous works such as~\cite{zela_understanding_2020} also indicate that the proposed results are often sub-optimal, especially when the search space is not well chosen. 
Specifically, since network weights are randomly initialized, promising operations can have poor initial weights such that the architecture optimization tends to entirely discard them. As a result, the practical relevance of DAS-proposed architectures depends heavily on network initialization as well as on training hyperparameters. Yet, in the context of large-scale computer vision problems such as image classification, a systematic analysis of DAS with respect to hyperparameter optimization is hardly affordable.

In this paper, we apply DAS to inverse problems with the main focus on the analysis of DAS w.r.t.~the impact of domain shifts, training hyperparameter choice and network initialization. Since signal recovery has not received nearly as much attention in the NAS literature as image classification, it allows to study a naive choice of parameters and settings without bias to known results and best practices. 
In the signal recovery setting, sequential architectures~\cite{zhang_beyond_2017} yield competitive results when learning to solve inverse problems, such that we can analyze the impact of the complexity of the search space more easily. 
Specifically, we compare the stability and sensitivity to hyperparameters of DAS-like architecture optimization in a simple, sequential search space as well as in a non-sequential search space, which we both propose, where the latter is inspired by the search space proposed in \cite{liu_darts_2019}.
We investigate two types of one-dimensional inverse problems which allow for extensive experiments 
for each setting in order to analyze the robustness of DAS. 
We show that DAS can automatically find well performing architectures, if the search space is well preconditioned.
Yet, our study also shows that the performance of DAS heavily depends on hyperparameter choices. 
Moreover, DAS shows a large variance for any set of hyperparameters, such that the suitability of parameters as well as the overall performance can only be judged when considering a large number of runs. This finding challenges the understanding of DAS as a one-shot method for NAS. 
Equally concerningly, we find that the estimated network performance using jointly optimized, shared weights is often not well correlated with the reconstruction ability of the final model after operation selection and re-training,~i.e.~the continuous relaxation in DAS seems to be quite loose. In particular, this makes the search for good hyperparameters by optimizing for the DAS training objective near-impossible. Hyperparameter optimization w.r.t.~the final architecture performance is even more expensive and seems to increase the variance in the results even further. 
Yet, overall, our study also shows that DAS can successfully be applied to inverse problems. Specifically, it improves over the competitive random search baseline by a significant margin, when the search space contains a variety of harmful and beneficial operations. This finding is crucial, since the search space can not always be assumed to be well preconditioned in novel applications.

\section{Differentiable Architecture Search}\label{DARTS}
We introduce DAS~\cite{liu_darts_2019} in a more detailed way in the following, as it is the basis of our analysis. While the originally proposed method optimizes so-called \textit{cells}, which are stacked in order to define the overall neural network architecture, and defines each cell in the form of a directed acyclic graph (DAG), we conduct most parts of our systematic study of the behavior of DAS on special sequential, and easy-to-interpret meta-architectures to be described below. To exclude that our findings are merely due to this special search space, we also consider experiments resembling the original DAS setup in~\cite{liu_darts_2019}.  
Our sequential architecture merely consists of $N$ nodes $x^{(i)}$, where $x^{(0)}$ represents the input data and the result $x^{(j+1)}$ of any layer is computed by applying some operation $o^{(j)}$ to the predecessor node $x^{(j)}$,~i.e.,
\begin{align}\label{eq:operation_summation}
    x^{(j+1)} = o^{(j)}(x^{(j)}, \theta^{(j)}), 
\end{align}
where $\theta^{(j)}$ are the (learnable) parameters of operation $o^{(j)}$. 
To determine which operation $o^{(j)}$ is most suitable to be applied to the feature $x^{(j)}$, one defines a set of candidate operations $o_t \in \mathcal{O}, t \in \{1,\dots, \mid \mathcal{O} \mid \eqqcolon T\}$ and searches over the continuous relaxation of Eq. \ref{eq:operation_summation} for
\begin{equation}
\label{eq:operationRelaxation}
        o^{(j)} = \sum_{t=1}^T \beta_{o_t}^{(j)} o_t, \qquad \beta_{o_t}^{(j)} = \frac{\mathrm{exp}(\alpha_{o_t}^{(j)})}{ \sum_{t'=1}^T \mathrm{exp}( \alpha_{o_{t'}}^{(j)})}
\end{equation}
where $\alpha = (\alpha_{o_t}^{(j)})$ are \textit{architecture parameters} that determine the selection of exactly one candidate operation in the limit of $\beta$ becoming binary. Instead of looking for binary parameters directly, the optimization is relaxed to the soft-max of continuous parameters $\alpha$. 

DAS formulates this search as a bi-level optimization problem in which both, the network parameters $\theta = \{ \theta^{(j)}\}_{j=1}^N$ and the architecture parameters $\alpha$, are jointly optimized on the training and validation set, respectively, via
\begin{align}
\label{eq:upperLevel}
    & \min_{\alpha} \mathcal{L}_{val}(\theta(\alpha), \alpha) \\
 \label{eq:lowerLevel}
  \text{s.t. }  & \theta(\alpha) \in \argmin_{\theta} \mathcal{L}_{\mathit{train}}(\theta, \alpha),
\end{align}
where $\mathcal{L}_{val}$ and $\mathcal{L}_{\mathit{train}}$ denote suitable loss functions for the validation and training data. The optimization is done by approximating \eqref{eq:lowerLevel} by one (or zero) iterations of gradient descent, and depends on several hyperparameters such as initial learning rates, learning rate schedules and weight decays for both architecture and model parameters.

At the end of the search, the discrete architecture is obtained by choosing the most likely operation $\hat{o}^{(j)} = \mathrm{argmax}_{o_t} \alpha_{o_t}^{(j)}$ for each node. Subsequently, the final network given by the architecture Eq. \ref{eq:operation_summation}, but using $\hat{o}^{(j)}$ instead of ${o}^{(j)}$, is retrained from scratch. Thus, the fundamental assumption that justifies the idea of DAS is that the performance reached by the final network architecture on the validation set (the \textit{architecture~validation}) is highly correlated with the performance of the relaxed DAS approach obtained in Eq. \ref{eq:upperLevel} (the \textit{one-shot~validation}). Only then, the architecture found during DAS optimization can also be expected to perform well after retraining. 
While previous works (to be summarized in the next section) have studied a search-to-evaluation gap,~i.e., the effect that the final network architecture's performance improves significantly by retraining from scratch, we further investigate whether this assumed correlation between \textit{one-shot~validation} and \textit{architecture~validation} is always given and in how far it depends on the choice of hyperparameters.

\section{Related Work}
In the last years neural architecture search (NAS) gained ever more interest due to its advantage over time-consuming manual trial-and-error network design. Especially DAS~\cite{liu_darts_2019} took a new step in the NAS area, as it was the first method to introduce gradient-based neural architecture search. By relaxing the operation choices in the network and thus allowing for gradient-based optimization, it shows significant advantages in the search of good neural architectures within only a few GPU days, resulting in finding architectures with impressive performances. Building on this pioneering work NAS research has gained significant momentum for further improvements over the original DAS approach 

\cite{Pham_2018,Liu_2018_progressivenas,dong_searching_2019,cai_proxylessnas_2019,Xie_2019_SNAS,Chen_2019_PDARTS,Akimoto_2019_Adaptive,XU_2020_PCdarts,He_2020_Milenas,Chen_2020_SDARTS,Wu_2021_SparseSupernet,zhang_2021_idarts}.
For example \cite{Chen_2019_PDARTS} presents an algorithm to progressively increase the depth of the searched architecture during training bridging the gap between search and evaluation performances. 
To narrow down the vast amount of literature building on the DAS approach, we focus here only on weight-sharing literature that is in line with our case study.

\noindent\textbf{Stability of DAS.\,\,}

There are only few works investigating 
the stability of DAS \cite{zela_understanding_2020,XU_2020_PCdarts,Chu_2020_FairDarts,Chen_2020_SDARTS}.
RobustDARTS~\cite{zela_understanding_2020} track the dominant eigenvalue $\lambda^{\alpha}_{\text{max}}$ of the Hessian during the architecture search and implement a regularization and early stopping criterion based on this quantity for a more robust DAS search. \cite{Chen_2020_SDARTS} picks up the relationship between the Hessian during the architecture search and the performance gap during search and evaluation time. They propose a perturbation-based regularization to smooth the validation loss landscape. 
\cite{XU_2020_PCdarts} find that only connecting partial channels into the operation selection leads to a regularized search to improve the stability.
\begin{figure}[t]
    \centering
    \includegraphics[width=0.99\columnwidth, height=2.5cm]{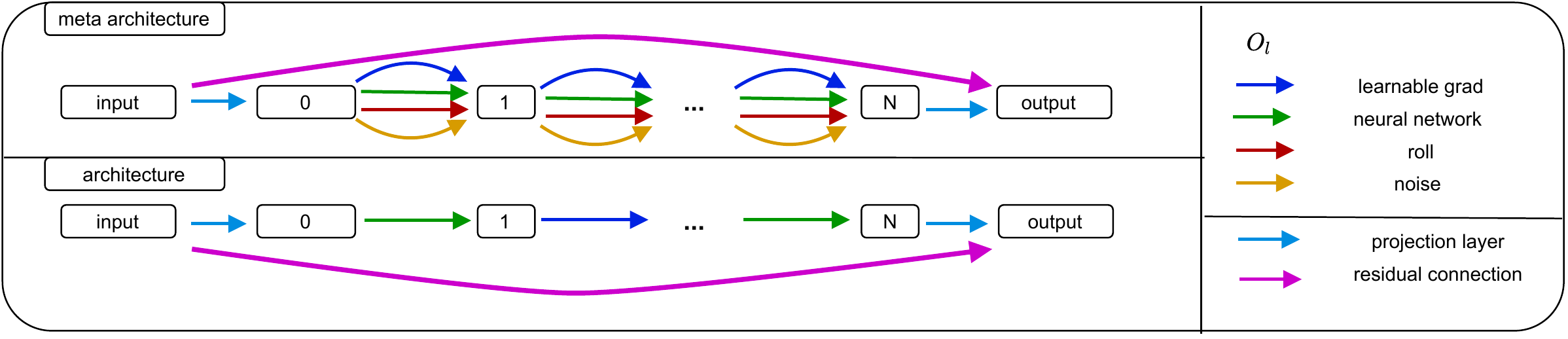}
    \caption{Investigated sequential meta-architecture. This setup is simple, yet it is able to represent DnCNN~\cite{zhang_beyond_2017}-like architectures. 
    \label{fig:arch}}
\end{figure} 
\cite{Chu_2020_FairDarts} use a sigmoid activation for the architecture weights instead of softmax to eliminate unfair optimization regarding the skip-connection operation.
\cite{yang_nas_2020} analyzes the contribution of each component in a NAS approach within the search space from \cite{liu_darts_2019}. They highlight that a performance-boosting training pipeline, often a result of expert knowledge, is more important for the evaluation of architectures than the search itself.
These findings motivate our analysis of the potential benefits of DAS in a different setting than image classification. 

\noindent\textbf{Reconstruction.\,\,}
Previous work on reconstruction of inverse problems via learned approaches has often focused on unrolled optimization schemes, such as unrolled PDHG in \cite{riegler_atgv-net:_2016} and \cite{adler_learned_2018}. These architectures, also referred to as variational networks \cite{klatzer_learning_2016,hammernik_deep_2017}, are constructed by  unrolling existing optimization routines that solve inverse problems and adding learning components in blocks which are either recurrent, as e.g.~in \cite{aggarwal_modl_2019} or fully independent as in \cite{hammernik_learning_2018}. In this investigation we will focus on parameterized gradient descent layers which can be seen as the most fundamental building block of these optimization routines.
\section{Proposed Search Spaces}
The significant advantages in computational efficiency over discrete architecture optimization methods along with the impressive performances of the final architectures have made DARTS and its variants highly attractive for automating the search for well-working neural networks. This framework itself is generic and thus applicable to any field of application, such as inverse problems.

Our following analysis of DAS for inverse problems will deliberately not be targeting settings that yield good results by design. In contrast, we propose two search spaces with different complexities that allow to analyze the stability and performance of DAS under varying degrees of difficulty, in ascending order:
\begin{itemize}
    \item finding a good (linear) sequence of operations from meaningful choices of operations,
    \item finding a good (linear) sequence of operations where the set of operations to choose from contains good operations as well as harmful operations (the model needs to learn to avoid these),
    \item finding a good non-linear, acyclic computational graph of operations from meaningful choices of operations (this is the conventional DAS setting),
    \item finding a good non-linear, acyclic computational graph of operations,  where the set of operations to choose from contains good operations as well as harmful operations (the model need to learn to avoid these).
\end{itemize} 
Such search spaces allow to investigate the properties of DAS methods under various and realistic conditions. Specifically, not for all tasks, we can assume that the set of well-performing, beneficial operations is given or even complete. In such setups, one would ideally want to be able to add new operation candidates to the search space and have the search determine which configuration will work best. Therefore, it is desirable that methods perform reliable even if poor operation choices are available.

\subsection{Sequential Search Space}
For the simpler, sequential search space, 
we propose the meta-architecture shown in Fig. \ref{fig:arch}, which should be specifically well-suited for examples of signal recovery from known data formation processes such as blurring and subsampling with noise. From a pre-defined set of operations, we choose operations sequentially before adding the output to a residual branch. Image recovery networks such as DnCNN~\cite{zhang_beyond_2017} are contained in this meta-architecture. In practice, we search for 10 successive layers. A detailed discussion of the proposed operations is given in Sec. \ref{sec:operations}. As discussed above, the search space deliberately contains benign as well as harmful operations. This allows the evaluation of the effectiveness of DAS in any setting via the distinction of two cases: Training on all operations versus training only on beneficial operations. \textit{A good architecture search algorithm should reliably find the optimal operations, even when presented with sub-optimal choices.}
\begin{figure}[t]
    \centering
    \includegraphics[height= 3.3cm, width=0.99\columnwidth]{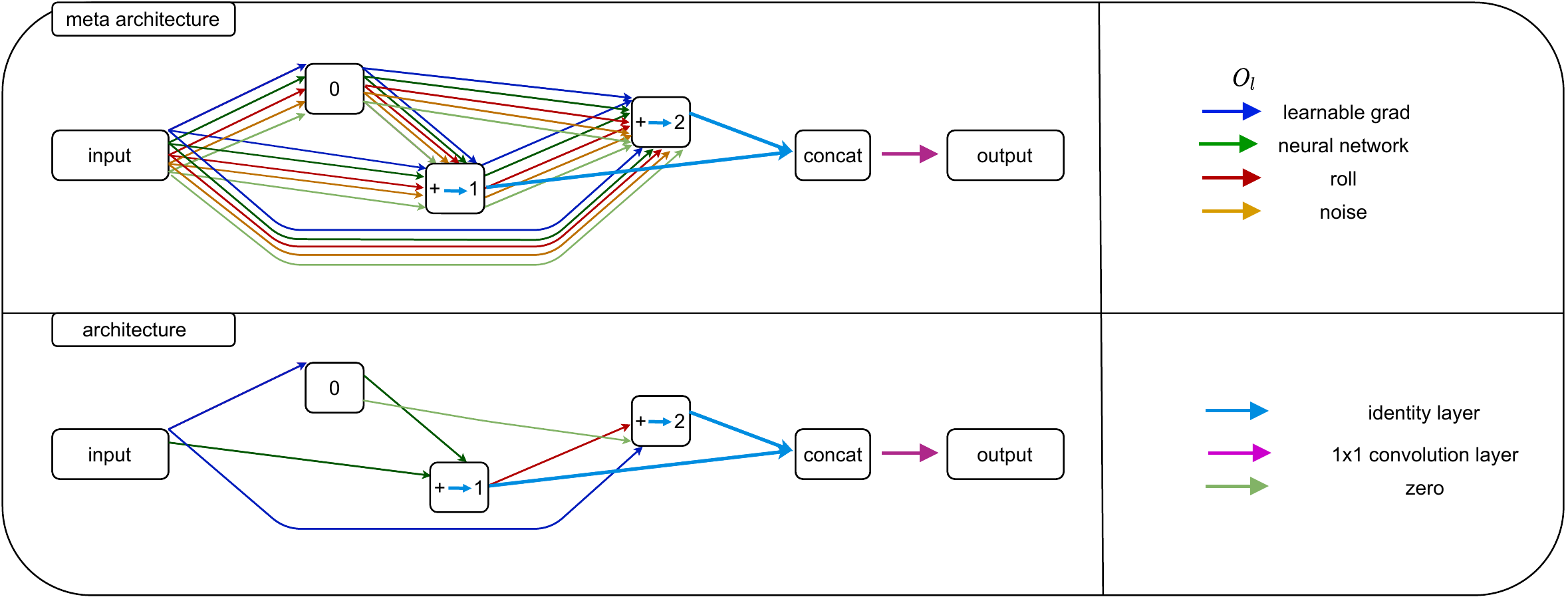}
    \caption{Investigated meta-architecture in the non-sequential search space.
    \label{fig:wide_arch}}
\end{figure} 
\subsection{Non-Sequential Search Space}\label{sec:exp_non_sequential}For the more complex, non-sequential search space, 
we construct a cell structure with 5 states, and allow for arbitrary forward connections among the same set of operations as in the sequential setting, but also allowing for a $\{\textit{zero}\}$ operation, resulting in up to $15$ operational connections. The output of the last two states is then concatenated and flattened via a 1D convolution. We utilize two of these cells in succession, so that the depth of this search space with in total $10$ nodes is comparable to the sequential search space from above.

Fig. \ref{fig:wide_arch} visualizes an exemplary cell meta architecture during architecture search and the found cell architecture, which is then retrained. We choose in each cell one operation out of several as a connection between each node in the cell. This setting is more directly comparable to the original DAS formulation \cite{liu_darts_2019}, which contains a cell structure with multiple possible connections between sequential states, allowing for a larger degree in freedom in combining computational results.
The selected architecture for retraining then takes the operation between each node with the highest probability.

\subsection{Network Operations for Inverse Problems}
\label{sec:operations}
In both the sequential and non-sequential setting, we search for the optimal architecture that can be defined using operations selected from a defined set $\mathcal{O}_l$. Specifically, we propose to use four operations, two of which are benign and potentially beneficial by design. 
The first benign operation is motivated from rolled-out-architectures (see e.g. \cite{lista,shrinkageFields,VariationalNetworks}) and tries to embed model-based knowledge about the recovery problems into the networks architecture. In this paper we consider problems which can be phrased as linear inverse problem, in which the quantity $x$ ought to be recovered from data $y=Ax+ \textrm{noise}$ for a linear operator $A$. While the precise type of algorithm is typically dictated by (smoothness) properties of the regularization, a partially parameterized network-based approach has a lot of freedom to choose from template layers based on the mentioned inverse problem $y=Ax + \textrm{noise}$
\begin{equation}\label{eq:inverse_objective}
    \argmin_x D(Au, f),
\end{equation}
where $D$ is a data formation term arising from the distribution of noise present, i.e.\ $D(v, f) = \frac{1}{2}||v-f||^2$ for Gaussian noise. This optimization objective yields templates such as a gradient descent layer:
    $u^{k+1} = u^k - \tau A^T\nabla_x D(Au^k, f),$
for input $u^k$ and output $u^{k+1}$ of a new layer. For suitable chosen $\tau$, the application of this layer is guaranteed to reduce the objective \eqref{eq:inverse_objective}.
The gradient layer can be turned into a learnable operation by introducing a learnable mapping $n(u, \theta)$ after the gradient step,
\begin{equation}
    u^{k+1} = u^k - \tau A^T\nabla_x D(Au^k, f) - \tau n(u^k, \theta),
\end{equation} 
as a \textbf{learnable gradient descent layer} in our operation set $\mathcal{O}_l$. The second benign layer is a fully-learned \textbf{neural network layer} in our operation set $\mathcal{O}_l$, that learns an appropriate mapping $n(u, \theta)$ without knowledge of the operator $A$:
\begin{equation}
    u^{k+1} = n \left(u^k, \theta \right),
\end{equation}

For both layers, the learnable mapping $n(u, \theta)$ is parametrized by a small convolutional network, consisting of a convolution layer, followed by batch normalization, ReLU and a second convolution layer.
These two layers, learnable gradient descent layer and neural network layer, are by design beneficial operations. In contrast to these beneficial layers we also include two negative operations to each operation set; a gradient layer with white Gaussian noise, \textbf{noise layer}, and a \textbf{roll layer}, which rolls the inputs in all dimensions. We provide additional details for these operations in the supp.~material. In total, we set $\mathcal{O}_l = \{\mathrm{learnable~gradient~descent},~\text{2-layer-CNN},~\mathrm{roll}, ~\mathrm{noise}\}$.

\section{Evaluating DAS for Inverse Problems}
In the following, we describe the experimental setting in which we evaluate DAS for inverse problems. Thereby, we focus on small problem instances to be able to evaluate the framework not once but in several runs such as to evaluate the statistics of the results. This setup also allows to gain insights on the dependence of DAS' performance on the chosen hyperparameters.
\subsection{Experimental Setup}\label{ex:exp_setup}
\noindent\textbf{Data Generation.}
For a fast synthetic test we generate one-dimensional data sampling cosine waves of varying magnitude, amplitude and offset, and search for models to recover these samples from distorted measurements. We consider two distortion processes with varying difficulty: First, Gaussian noise and blurring and second, in addition to these, a subsampling by a factor of 4.

We generate these synthetic one-dimensional cosine data from $N=50$ equally spaced points $\omega_i$ on the interval $[-\frac{\pi}{2}, \frac{\pi}{2}]$ with the model
\begin{equation*}
    x_i = \cos(f \omega_i + O_x) + O_y 
\end{equation*}
for a random frequency $f$ drawn uniformly from the interval $[0, 2\pi]$ and offsets $O_x$ and $O_y$ drawn from a normal Gaussian distribution. Such random drawn waves comprise our ground truth training data. We then generate measured data via the linear operation $A$ and addition of noise,
\begin{equation*}
    y = Ax + n,\qquad n \in \mathcal{N}(0, \sigma_n).
\end{equation*}
These pairs $(y, x)$ represent our training data. We sample new examples on-the-fly during both training and validation, so that no confounding effects of dataset size exist. All validation and training loss evaluations are each based on 2432 randomly drawn samples. The performance of all models is evaluated in terms of their average peak signal to noise ratio~(PSNR) on validation data. 
For all experiments we chose $\sigma_n=0.01$. For the \textit{blur} experiments, the linear operator $A$ is a Gaussian blur with kernel size $7$ and $\sigma_b=0.2$. For the \textit{downsampling} experiments, this Gaussian blur is followed by a subsampling by a factor 4.

\noindent\textbf{Hyperparameter Optimization.}
Our one-dimensional case study allows us to optimize DAS training hyperparameters with more granularity than it would be possible for image classification tasks. While we run our first experiments using manually tuned hyperparameters (see Appendix for details), we also consider the behavior and stability of DAS under optimized hyperparameters. We stress that we consider this mainly as an analysis tool - given that NAS itself is a hyperparameter optimization on which we stack another, and acknowledging that this optimization is practically intractable when larger problems are considered. To improve hyperparameters, we apply BOHB~\cite{falkner_bohb_2018}, a Bayesian optimization method with hyperband \cite{li_hyperband_2018} and run BOHB for 128 hyperband iterations, which is an affordable budget in this one-dimensional data setting. It is important to note that BOHB is not an exhaustive search and thus there are no guarantees for success within our budget or even in general in a way, that it finds the globally optimal hyperparameters for the just mentioned objective. The usage of BOHB as such covers the problem of hyperparameter optimization partially, but in general there is no simple fix of DAS via hyperparameter search - which is itself a notable statement about the algorithm. 
In particular we optimize the hyperparameters with respect to first the one-shot validation performance, ``BOHB-one-shot'', and second the final architecture performance, ``BOHB''. Note here that hyperparameter search that maximizes the final architecture performance instead of the one-shot validation performance is twice as expensive (on top of the already expensive hyperparameter search), due to the need for retraining. 

\subsection{Results}
\label{sec:results}
We first investigate the performance of DAS on the simplified sequential search space to validate the one-shot search property of DAS. For our analysis, we are not only interested in the best found architecture but also in the statistics of the search to leverage the advantages of the proposed, efficient setup. We therefore evaluate 75 trials of DAS as well as baselines such as setting all operations to \emph{Learnable Grad.} or \emph{Net} (i.e.~learnable gradient descent or 2-layer CNN), picking a random architecture as well as performing a random search within an equal time budget as required by a DAS run. We summarize the results in Tab \ref{tab:itLooksGood}.  
The results indicate that DAS works well for inverse problems. It proposes successful architectures given the complete operation set $\mathcal{O}_l$ for both considered data formation methods, \emph{blur} and \emph{downsampling}. Thereby, it outperforms architectures consisting of only one good operation. Practically, these experiments thus lead to a first interesting result for applied inverse problems: The best found architecture is a hybrid version  that mixes both beneficial operations, possibly suggesting that the best way to approach inverse problems are neither plain (convolutional) networks nor pure unrolling schemes.

Next, we compare DAS to a random search approach (random selection of the operation at each layer). 
To allow for a comparison at an equal time budget (Random search in Tab. \ref{tab:itLooksGood}), we evaluate 5 randomly selected architectures and report the best. One random evaluation, i.e.~the training of one random architecture takes on avg.~57 sec.~versus 2min.~39 sec for one DAS run. For the sequential search space that purely consist of benign operations, random search outperforms DAS with a median PSNR of $22.75$ versus $21.6$ on \emph{blur} and $17.56$ versus $16.66$ on \emph{downsampling}. Thus in this scenario, \emph{random search outperforms DAS when used as a one-shot model}. 
This is different when harmful operations are added. For a search on the full operations set $\mathcal{O}_l$, DAS can clearly outperform this simple random baseline. 

For further analysis, we additionally investigate how many random search runs are needed, to improve over the DAS median for the set of all operations: Random search needs on average (10 runs) 49 random search steps to improve over the DAS median of $18.57$ PSNR. 
While this observation is overall motivating, we also observe that the performance of DAS significantly drops on average as well as in the median when all operations are considered (compared to only using benign operations). Especially on the blurred data the PSNR drops in the median from 21.6 (good ops.) to 18.57 (all ops.). This effect is undesired: ideally, DAS should be able to reliably filter out harmful operations. 

\begin{table*}
    \centering
    \caption{Architecture validation PSNR values for 1D inverse problems. Shown is the maximal, mean and median PSNR over 75 trials.    \label{tab:itLooksGood}}
\resizebox{\textwidth}{!}{
    \begin{tabular}{c@{\hspace{0.1cm}}|@{\hspace{0.1cm}}c@{\hspace{0.1cm}}|@{\hspace{0.4cm}}c@{\hspace{0.4cm}}c@{\hspace{0.4cm}}c@{\hspace{0.4cm}}|@{\hspace{0.4cm}}c@{\hspace{0.4cm}}c@{\hspace{0.4cm}}c@{\hspace{0.3cm}}|c}
    \toprule
    &Data  &  \multicolumn{3}{c@{\hspace{0.4cm}}|@{\hspace{0.4cm}}}{Blur} &  \multicolumn{3}{c@{\hspace{0.3cm}}|}{Downsampling}
        &\\
        \midrule
        &Method  &  \multicolumn{3}{c@{\hspace{0.4cm}}|@{\hspace{0.4cm}}}{\hspace{-0.3cm}Architecture Val. (PSNR)} &  \multicolumn{3}{c@{\hspace{0.3cm}}|}{\hspace{-0.3cm}Architecture Val. (PSNR)}
        &Runtime
        \\
        & & Max. & Mean & Med. & Max. & Mean & Med.&sec.\\
        \midrule
    \multirow{4}{*}{\rotatebox{90}{good ops.}}
    &Learnable Grad. only &  17.45 &  16.36 & 16.49 &   14.35 &  13.24 & 13.55& 0:57 \\
    &Nets only &   21.63 &  19.45 & 20.71 &  16.92 &  13.13 & 14.05&0:57\\  
    \cmidrule{2-9}
    &DAS    &  23.46 &  21.56 & 21.60&   18.03 &  16.36 & 16.66& 2:39\\
    &Random   & 24.04 & 22.00 & 22.16 &18.10  & 16.74  & 16.78 &0:35 \\
    &Random Search   & \textbf{24.04} & \textbf{22.85} &  \textbf{22.75}&   \textbf{18.10}  & \textbf{17.62} & \textbf{17.56} &2:55 \\
    \cmidrule{1-9}
    \multirow{2}{*}{\rotatebox{90}{all}}
    &DAS   & \textbf{22.86} &  \textbf{15.64} & \textbf{18.57}&\textbf{18.01} &  \textbf{15.39} & \textbf{16.12}&2:53 \\
    &Random  &  20.86 &  9.45 & 8.10&13.78 &  5.08 & 4.31&0:35\\
    &Random Search &  20.86 &  12.39 & 12.34& 13.78 &  9.61 & 9.77&2:55\\
    \bottomrule
    \end{tabular}
}
    \vspace{0.1cm}

\end{table*}

Since the original DAS formulation  in \cite{liu_darts_2019} contains a cell structure with multiple possible connections between sequential states, allowing for a larger degree of freedom in combining computational results, it is a-priori conceivable that some of the stability of DAS could be conferred through this structure. Therefore, we now analyze the DAS performance on the non-sequential DAS like search space exemplified in Fig. \ref{fig:wide_arch}. Table \ref{tab:non_sequential_h1} however shows that this wider search space does not improve the overall performance. Indeed the non-sequential search space hampers not only the DAS search significantly but also all other approaches, resulting in lower architecture performances for both data formations. In this setup, the \emph{Nets only} baseline, that uses the 2-layer CNN for all operations, performs best. As above, we observe a significant drop in the performance of DAS when harmful operations are included in the search space. In this case, as before, DAS can significantly outperform the random baseline but not reliably determine the obviously best operation. As the performance in this non-sequential search space is lower than in the sequential search space, we consider only the latter one in the following.
From a theoretical perspective, we argue that DAS should be able to determine which operations are harmful: If we assume that the validation accuracy during the optimization correlates with the validation accuracy of the final architecture, harmful operations should be excluded early on in the optimization process. Therefore, in the following section, we study this correlation and investigate whether the behavior can be improved by optimizing training hyperparameters.

\begin{table}
    \centering
     \caption{Architecture validation PSNR values for 1D inverse problems for the non-sequential search space. Shown is the maximal, mean and median PSNR over 100 trials.\label{tab:non_sequential_h1}}
    \resizebox{\columnwidth}{!}{  
\begin{tabular}{c@{\hspace{0.1cm}}|@{\hspace{0.1cm}}c@{\hspace{0.1cm}}|@{\hspace{0.4cm}}c@{\hspace{0.4cm}}c@{\hspace{0.4cm}}c@{\hspace{0.4cm}}|@{\hspace{0.4cm}}c@{\hspace{0.4cm}}c@{\hspace{0.4cm}}c@{}}
    \toprule
    &Data  &  \multicolumn{3}{c@{\hspace{0.4cm}}|@{\hspace{0.4cm}}}{Blur} &  \multicolumn{3}{c}{Downsampling}
        \\
        \midrule
        &Method  &  \multicolumn{3}{c@{\hspace{0.4cm}}|@{\hspace{0.4cm}}}{\hspace{-0.3cm}Architecture Val. (PSNR)} &  \multicolumn{3}{c}{\hspace{-0.3cm}Architecture Val. (PSNR)}
        \\
        & & Max. & Mean & Med.&  Max. & Mean & Med.\\
        \midrule
\multirow{4}{*}{\rotatebox{90}{good ops.}}
&Learnable Grad. only &  13.19 &  12.41 & 12.38  &  11.30 & 8.89 & 9.59\\
&Nets only &  \textbf{16.35} & \textbf{14.83}  & \textbf{15.82}  & \textbf{13.63} & \textbf{13.07} & \textbf{13.06}\\  
\cmidrule{2-8}
&DAS   &  15.34 &  13.08 & 12.51 &13.22 & 10.22 & 10.43\\
&Random  & 16.20 & 11.29 & 11.88    &13.15 & 6.26 & 5.72\\
\cmidrule{1-8}
\multirow{2}{*}{\rotatebox{90}{all}}
&DAS   &  \textbf{16.15} & \textbf{13.56}  & \textbf{13.73}  &13.31 & \textbf{9.47} & \textbf{8.61}\\
&Random  &  16.05 & 9.56  & 8.17  &\textbf{13.39} & 4.37 & 3.21\\
\bottomrule
    \end{tabular}}
    \vspace{0.1cm}

\end{table}

\subsection{Architecture and DAS Performance}\label{sec:correlation} 
\begin{figure}[ht]
    \centering
    \includegraphics[width=0.48\textwidth]{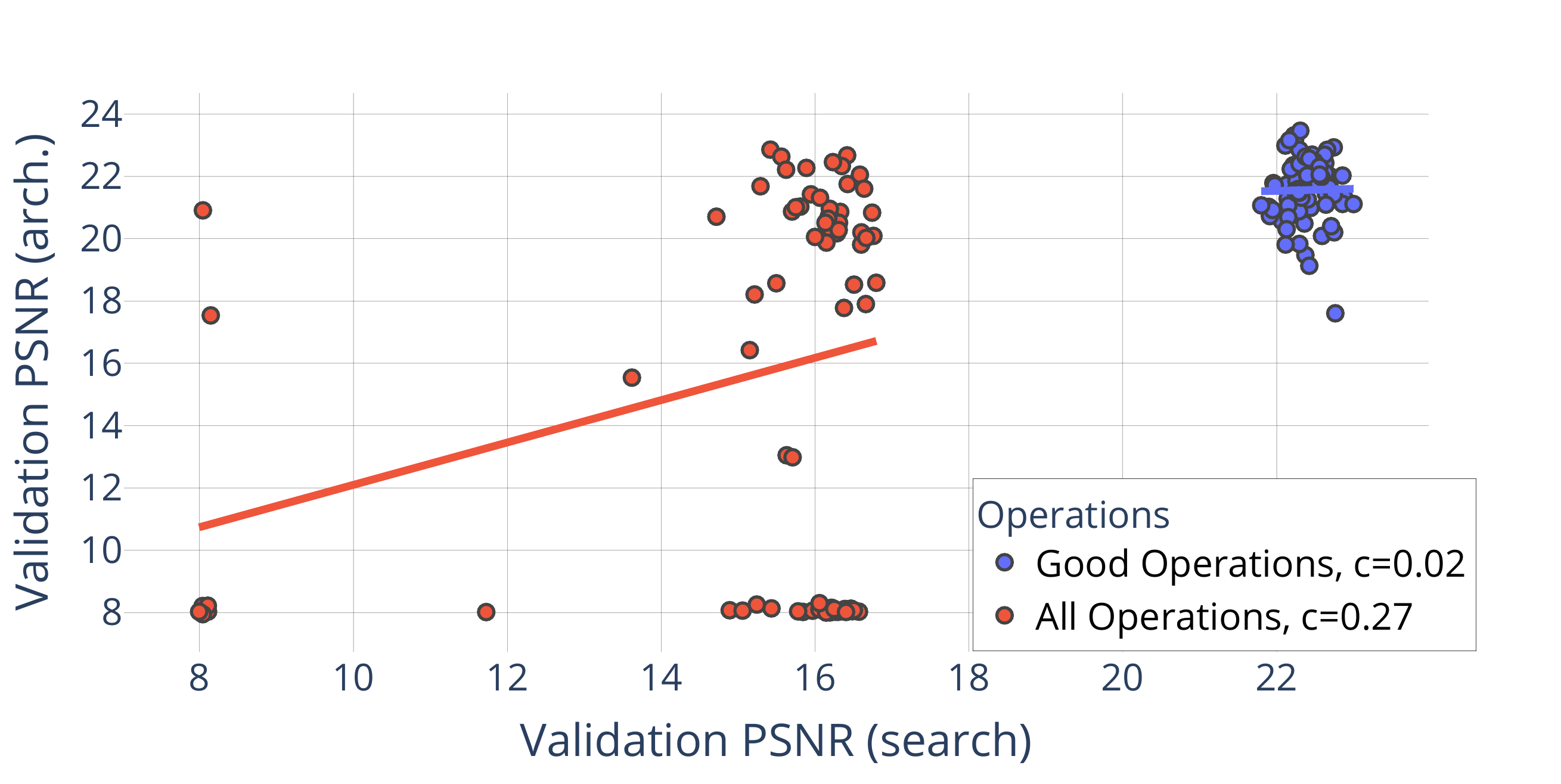}
    \includegraphics[width=0.48\textwidth]{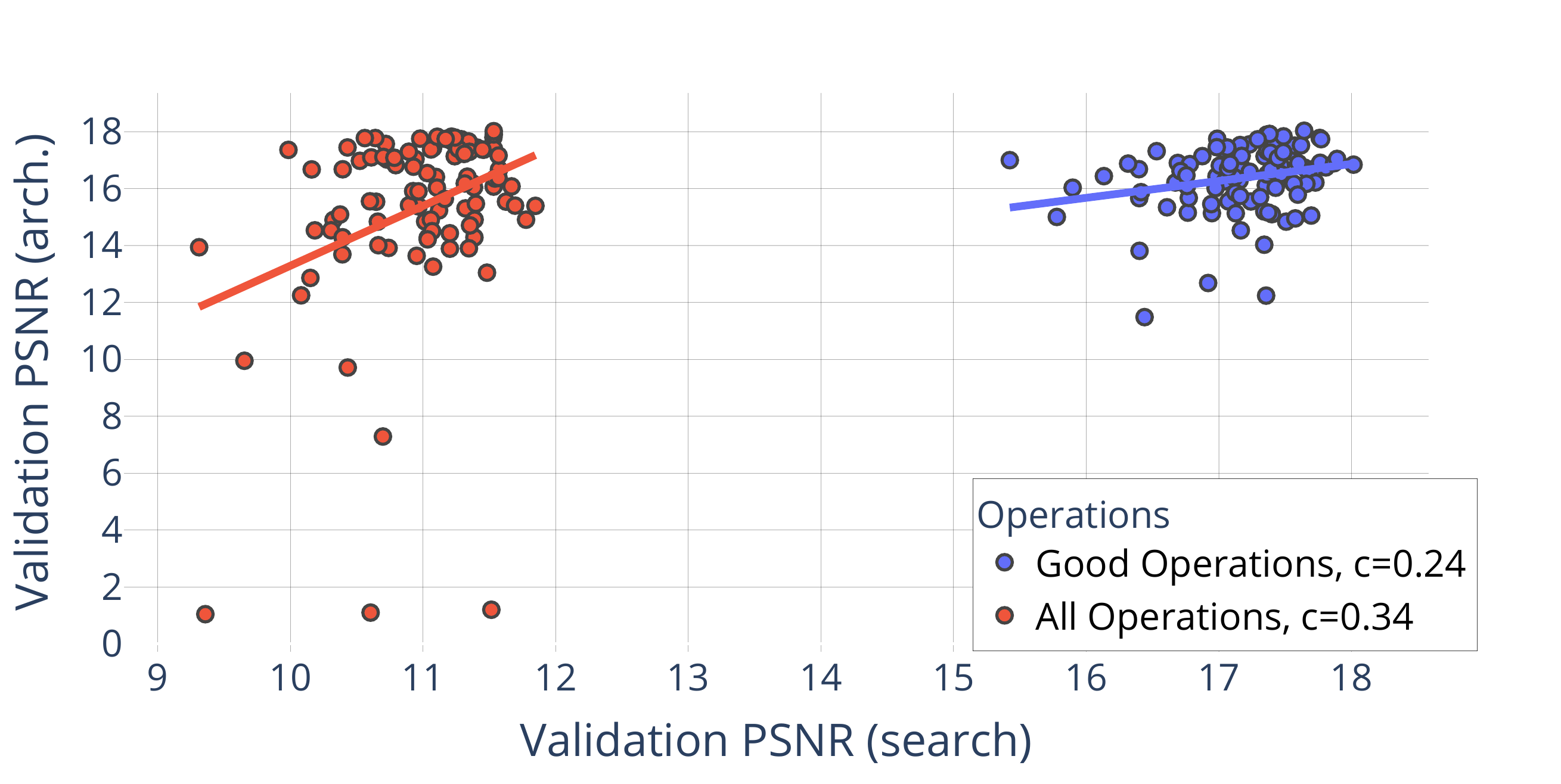}
    \caption{Scatter plot corresponding to Tab. \ref{tab:itLooksGood} showing architecture PSNR (y-axis) plotted against 1-shot validation PSNR (i.e. the validation performance on the DAS objective). Left: Blur. Right: Downsampling.
    \label{fig:correlated}}
\end{figure}

Figure \ref{fig:correlated} takes a closer look at the trials considered in Tab.~\ref{tab:itLooksGood}, scattering the values of all trials separately with architecture performance (y-axis), which is computed after retraining the final architecture versus the direct validation performance of the one-shot architecture (x-axis). We also plot a regression line over all trials and report the correlation of all trials in the legend, showing the linear fit has limited expressiveness. As already discussed, the correlation of these quantities is a fundamental assumption of DAS. However, this first experiment indicates a correlation problem: The assumption that a better \textit{one-shot~validation} implies a better \textit{architecture~validation} does not always seem to be true. 

These plots also show that DAS' behavior is highly problem-dependent: The \emph{downsampling} dataset (bottom), shows that, although the mean value of DAS can be non-optimal, search performance and architecture performance are weakly correlated, even if the best architecture only has average search performance. The closely related \emph{blur} dataset (top) shows an entirely different behavior with different ``failure'' cases, from which we can observe with the given hyperparameters 
that 1) either DAS proposes architectures with low (one-shot) search validation PSNR (i.e.~it fails), or that 2) DAS works but does not predict a useful architecture (low architecture validation PSNR), or that 3) DAS does predict a useful architecture, but is unrelated to its search performance. Only the best proposed architectures perform well in both. To further analyze the correlation, we investigate DAS behavior with different training hyperparameters. 
\begin{table}[t]
        \caption{Architecture validation PSNR values for 1D inverse problems with different hyperparameter. Shown is the maximal, mean and median PSNR over 75 trials.
        \label{tab:tabular_results}
        }
    \resizebox{\columnwidth}{!}{  
    \begin{tabular}{c@{\hspace{0.1cm}}|@{\hspace{0.1cm}}c@{\hspace{0.1cm}}|@{\hspace{0.3cm}}c@{\hspace{0.3cm}}c@{\hspace{0.3cm}}c@{\hspace{0.3cm}}|@{\hspace{0.3cm}}c@{\hspace{0.3cm}}c@{\hspace{0.3cm}}c}
    \toprule
        Data & Hyperparameters & \multicolumn{6}{c}{Architecture Validation (PSNR)}\\
        & & \multicolumn{3}{c@{\hspace{0.3cm}}|@{\hspace{0.3cm}}}{Good Ops.} & \multicolumn{3}{c}{All} \\
        & & Max. & Mean & Med. & Max. & Mean & Med.\\
        \midrule
        \multirow{5}{*}{Blur} & H1 & 23.46 & 21.56 &  21.60 & 22.86 & 15.64 & 18.57 \\
         & H2 & 23.46 & 21.43 &  21.63 & \textbf{23.10} & \textbf{16.77} & \textbf{19.88} \\
         & BOHB-one-shot-Blur & 22.83 & 20.86 &  20.75 & 22.47 & 15.57 & 18.04 \\
         & BOHB-one-shot-DS & 22.33 & 20.65 &  20.96 & 22.41 & 14.43 & 14.41 \\
         & BOHB-Blur & \textbf{23.57} & \textbf{22.05} &  \textbf{22.38} & 22.94 & 12.76 & 8.21 \\
         \midrule
        \multirow{5}{*}{
        Downsampling} & H1 & 18.03 & 16.36 &  16.66 & 18.01 & 15.39 & 16.12 \\
         & H2 & 18.20 & 16.57 &  16.78 & 17.82 & \textbf{15.93} & \textbf{16.21} \\
         & BOHB-one-shot-Blur & \textbf{18.42} & \textbf{16.83} &  \textbf{16.95} & 17.73 & 14.36 & 14.57 \\
         & BOHB-one-shot-DS & 17.51 & 15.33 &  15.84 & \textbf{18.12} & 12.36 & 13.65 \\
         & BOHB-Blur & 18.26 & 14.63 &  15.93 & 17.91 & 15.04 & 15.44 \\
        \bottomrule
    \end{tabular}
    }
    \vspace{0.1cm}

\end{table}

\begin{figure}[ht]
    \centering
    \includegraphics[width=0.48\textwidth]{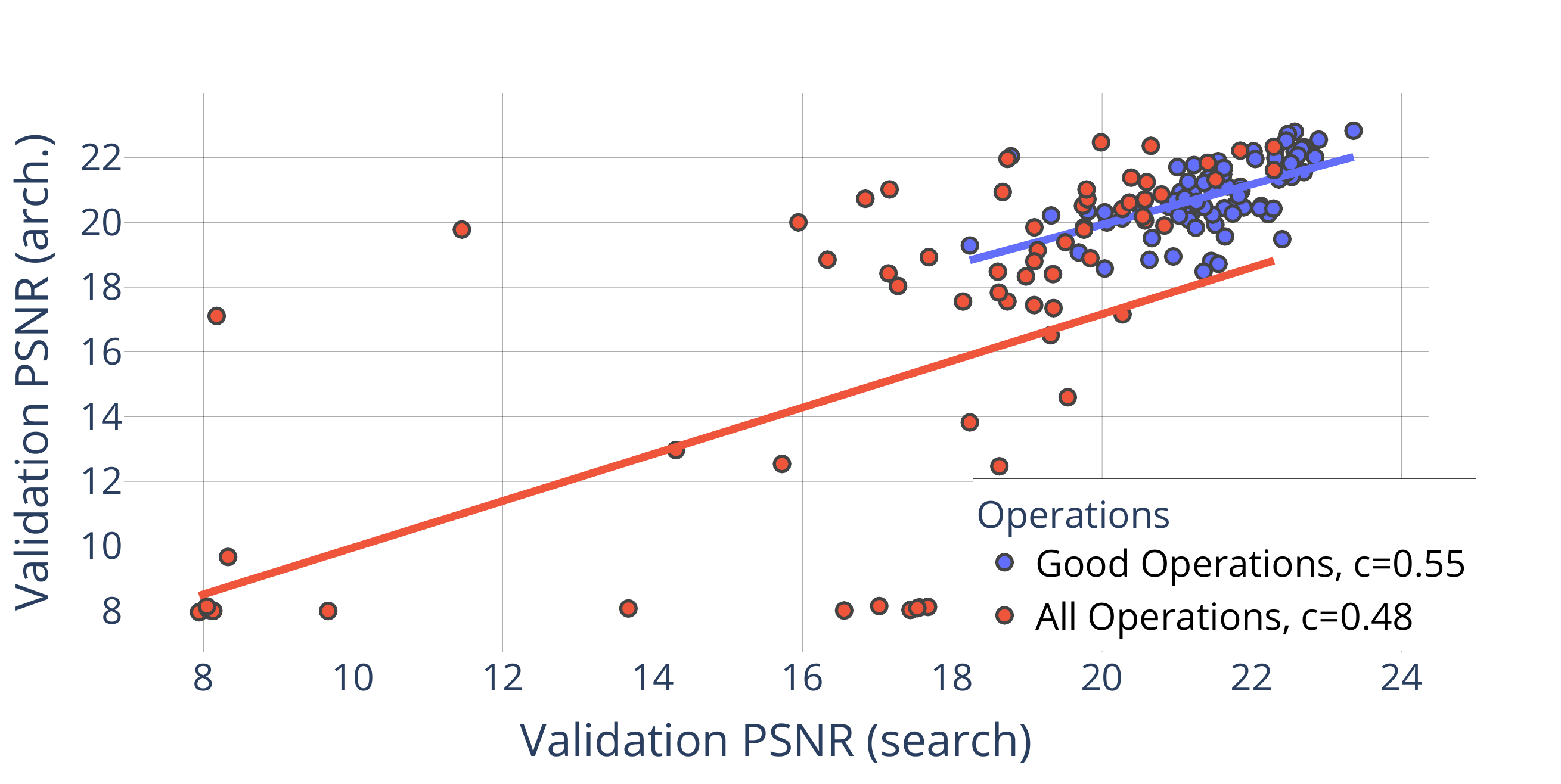}
    \includegraphics[width=0.48\textwidth]{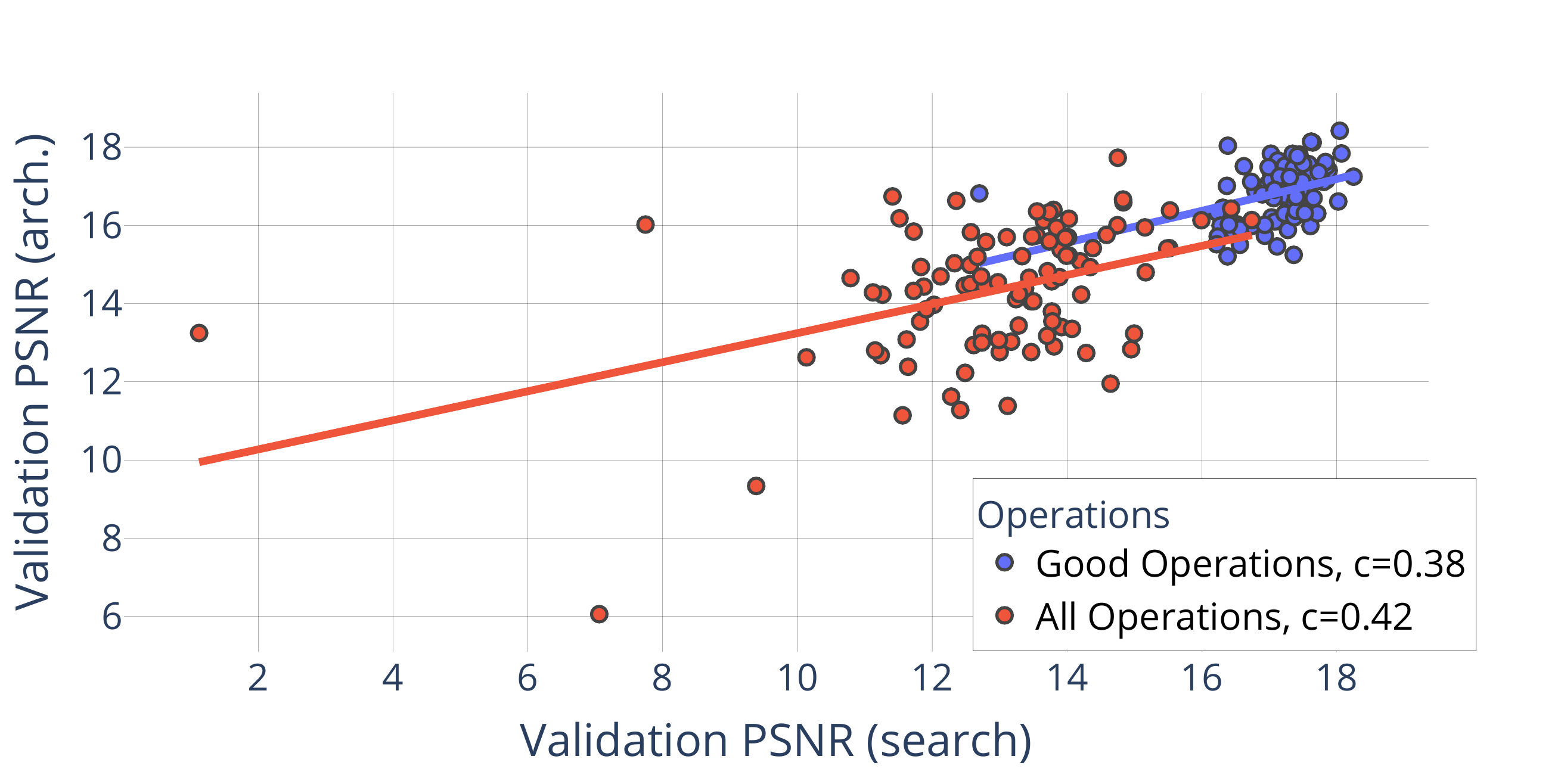}
    \includegraphics[width=0.48\textwidth]{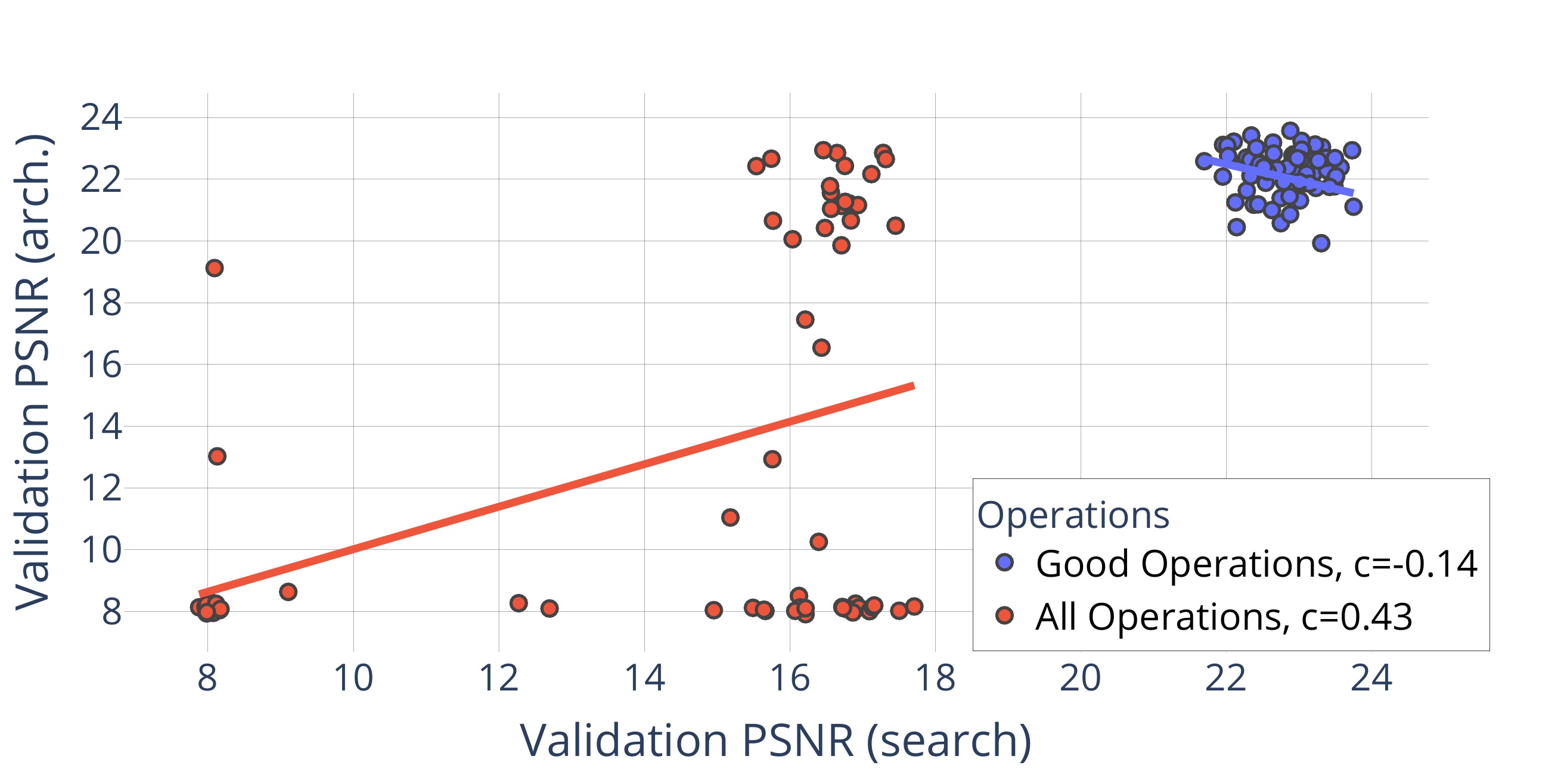}
\includegraphics[width=0.48\textwidth]{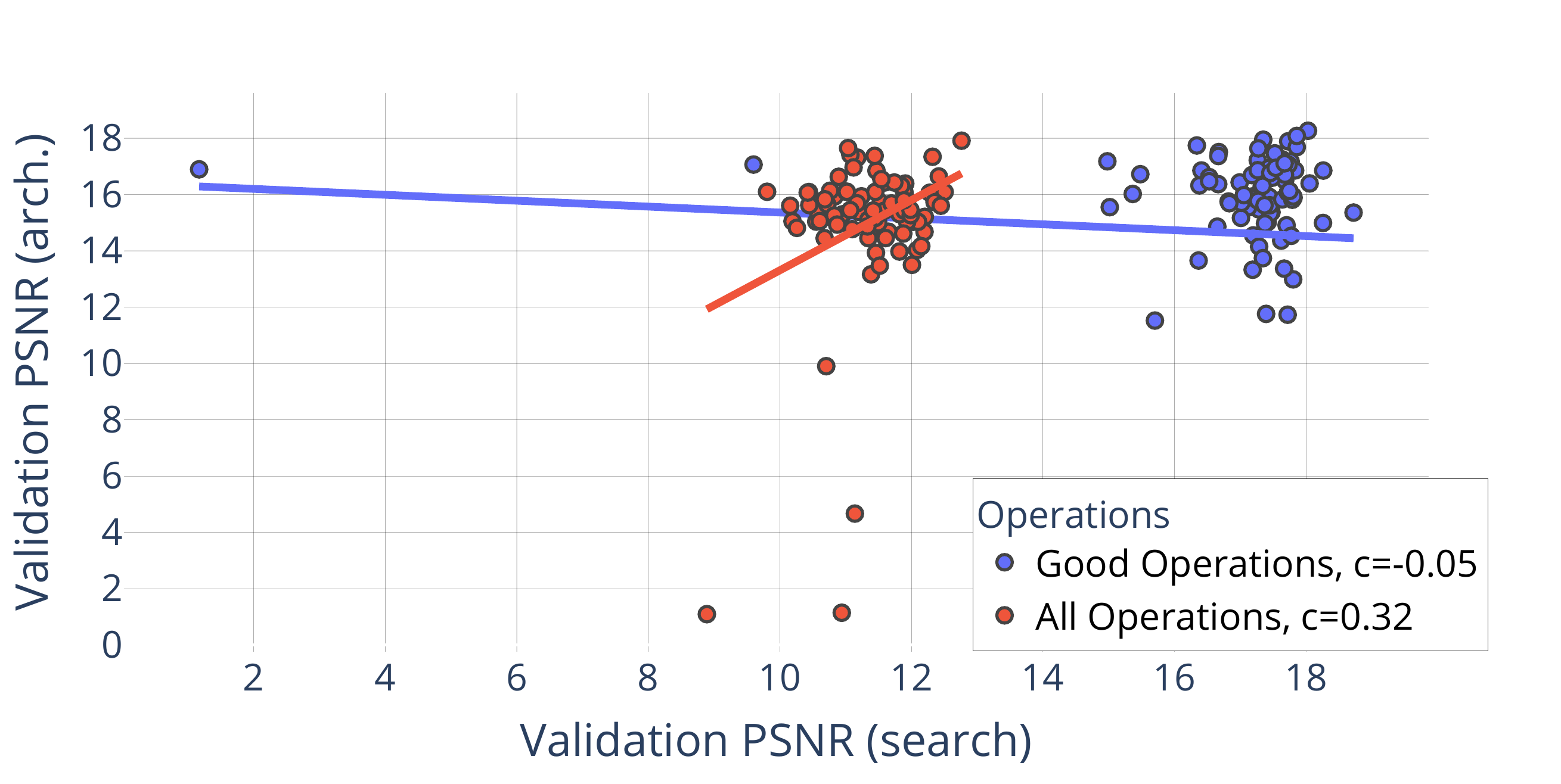}
    \caption{Scatter plot corresponding to Tab. \ref{tab:tabular_results} with BOHB-optimized hyperparameters, showing architecture PSNR (y-axis) plotted against one-shot validation PSNR (x-axis). Top (left): Blur with hyperparameters  \textit{BOHB-one-shot-Blur}. Top(Right): Downsampling  \textit{BOHB-one-shot-DS}. Bottom (Left): Blur with hyperparameters  \textit{BOHB-Blur}. Bottom (Right): Downsampling with hyperparameters  \textit{BOHB-Blur}.
    \label{fig:correlated2}}
\end{figure}

We evaluate DAS using 5 different training hyperparameter set; two are chosen manually, H1 and H2 (H1 are the hyperparameters used in Sec. \ref{sec:results}), whereas the other three are tuned using BOHB, as descibed in Sec. \ref{ex:exp_setup}. We use BOHB to tune hyperparameters for the one-shot validation performance for both \emph{blur} (BOHB-one-shot-Blur) and \emph{downsampling} (BOHB-one-shot-DS), individually, and also to target the final validation performance for \emph{blur} (BOHB-Blur). The DAS search results for different training hyperparameters are given in Tab. \ref{tab:tabular_results}. For additional visualization, we plot the results for all BOHB found hyperparameter trials in Fig. \ref{fig:correlated2}. The plot shows that the correlation for both data formation methods increases with the corresponding BOHB-one-shot tuned hyperparameters, with also a higher range of the search validation PSNR, as expected. This experiment also shows a rather surprising outcome: In the case of \emph{blur}, the average performance is on par with the manually chosen hyperparameters H1 and H2, whereas the performance for \emph{downsampling} decreases, especially when all operations are considered. In addition, the best architecture PSNR over 75 trials decreases in both cases. Overall, the apparent stabilization via optimization of the search loss removes not only negative, but also positive outliers. Furthermore, hyperparameters optimized for one dataset do not transfer well to the other. 
Using BOHB to target the final validation performance for \emph{blur} (BOHB-Blur) instead of the one-shot validation performance has also a positive impact on the one-shot validation and architecture validation correlation, compared to the manually chosen hyperparameters H1 and H2 in Fig. \ref{fig:correlated}, but not to the same amount as for the BOHB-one-shot hyperparameters. However, these hyperparameters successfully increase the max.~architecture performance. 
Overall, hyperparameters optimized with BOHB on the one-shot validation have to be considered with caution. This can be seen by cross-checking their performance, i.e.~evaluating the BOHB-one-shot-Blur hyperparameters for Downsampling and the BOHB-DS hyperparameters for \emph{blur}.
For the case of \emph{blur} and \emph{all operations} in Tab. \ref{tab:tabular_results}, the dedicated BOHB-one-shot-Blur hyperparameters are significantly more stable (measuring median PSNR) than the BOHB-one-shot-DS hyperparameters, although their maximal PSNR is very close.
When changing the domain to \emph{downsampling}, the exact opposite holds: BOHB-one-shot-Blur hyperparameters improve over BOHB-one-shot-DS hyperparameters in terms of stability. Note that this could be due to both, the missing correlation between one-shot and architecture validation as well as the missing guarantee of any Bayesian search to find the optimal hyperparameters.
In addition, Tab. \ref{tab:tabular_results} even demonstrates that the manual hyperparameters H1 and H2 lead to a better average performance compared with BOHB tuned hyperparameters.

In conclusion we find two schools of thought when evaluating the performance of DAS. For maximal performance, we should \emph{not understand DAS as a one-shot search approach}, but as a component in a \emph{larger search that proposes trial architectures}. For average performance, and immediate performance with a single DAS run, we should be optimizing the search performance and maximize its correlation with architecture performance - although as our experiments show, this is non-trivial even when searching for these hyperparameters in an automated fashion.
We stress that the two directions are not at odds with each other, yet problems can arise in the literature when comparing proposed improvements of DAS across both. Some algorithmic improvements of DAS are more likely to improve best-case performance, whereas others are more likely to impact single trial stability. If these two are not carefully compared, then best-case results, which do provide better benchmark numbers, can appear to supersede stability results.
Here, we discuss this effect for a simplified case study, but for large-scale DAS in image classification, where trials are expensive and fixed random seeds are tempting,
such a dichotomy makes it fairly difficult to evaluate and classify the manifold improvements of DAS.

\subsection{Improving the Initialization}
\begin{figure}[ht]
    \centering
    \includegraphics[width=0.48\textwidth]{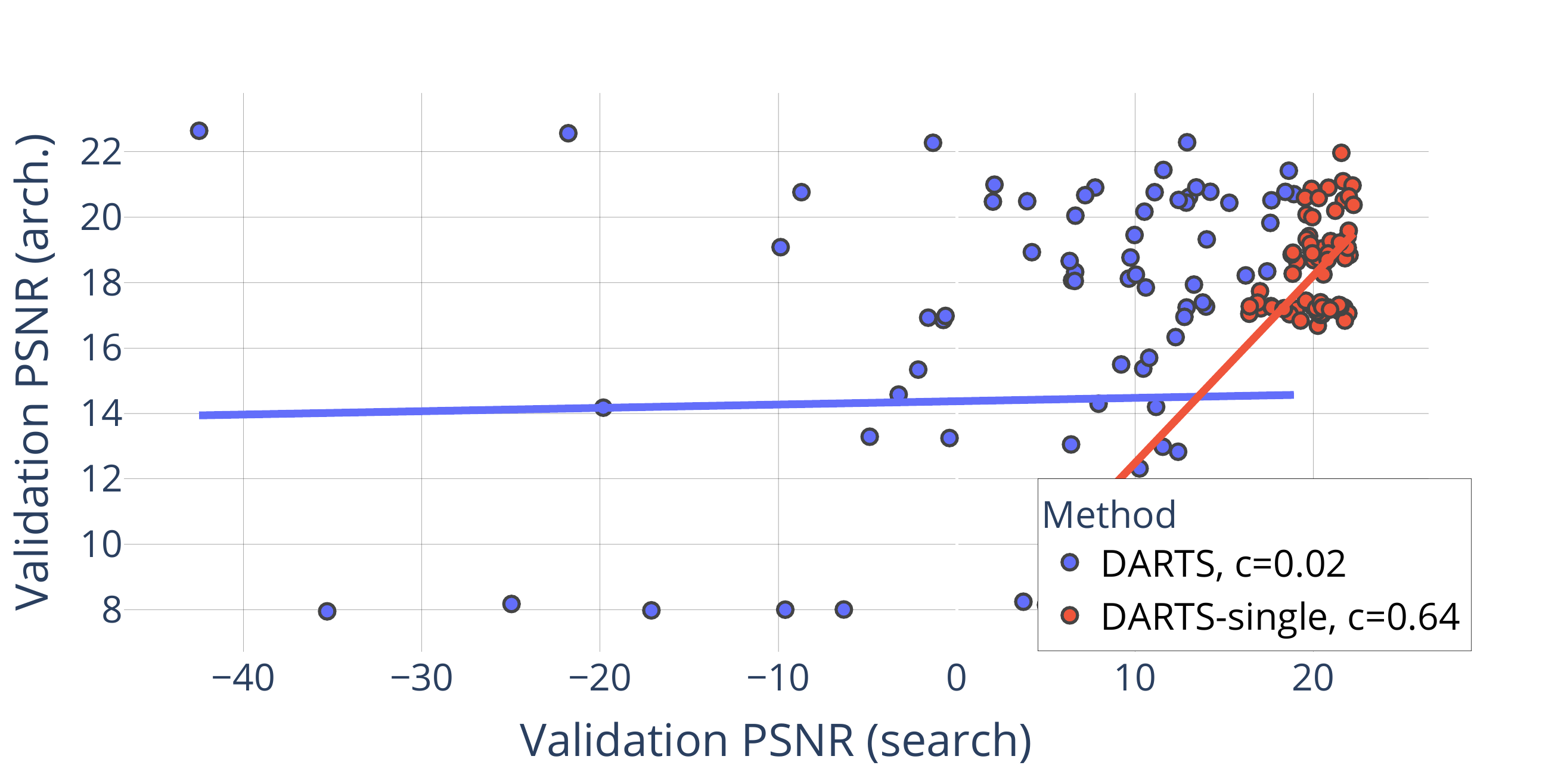}
    \includegraphics[width=0.48\textwidth]{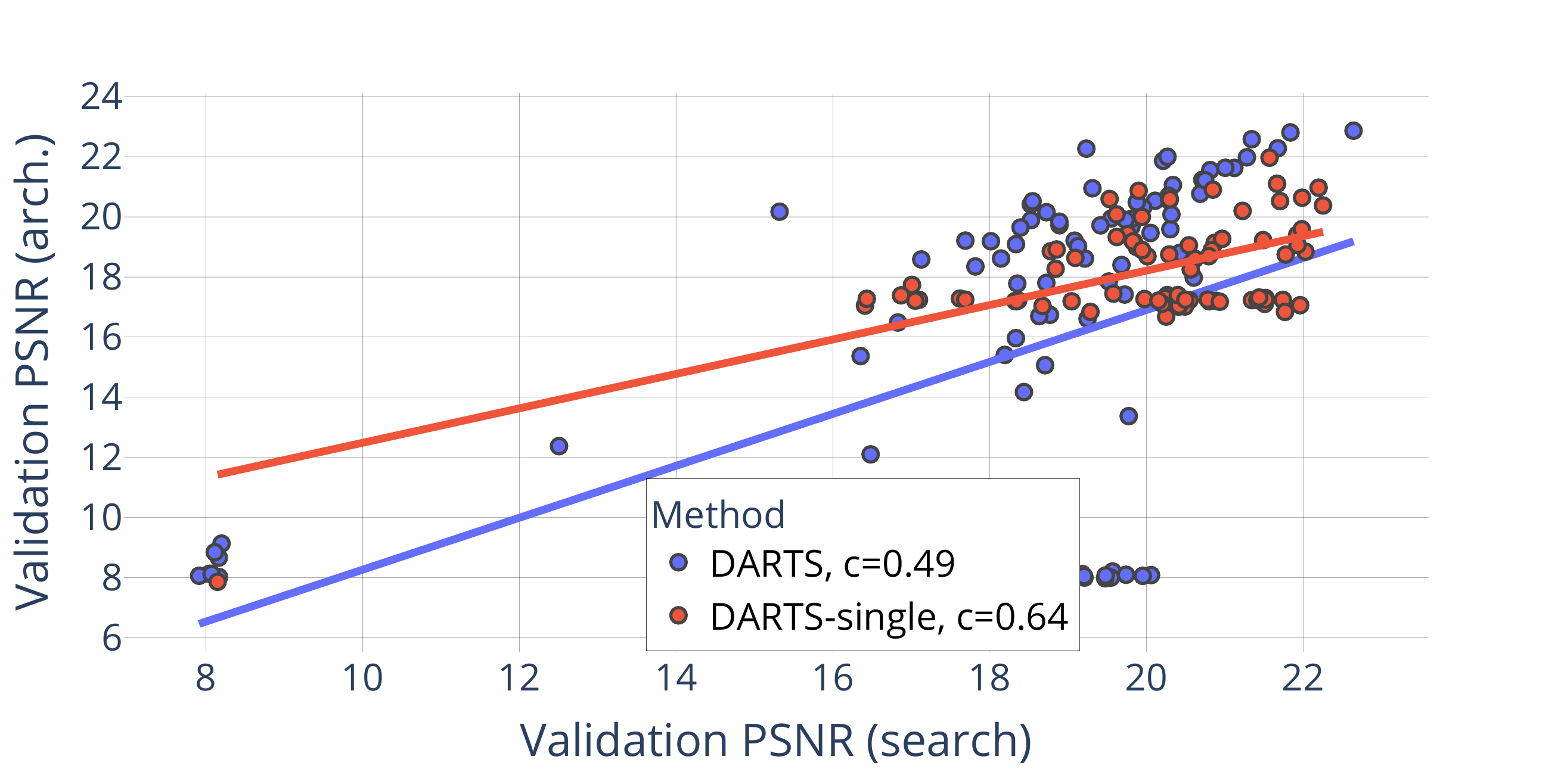}
    \caption{Left: Architecture search with BOHB-searched hyperparameters for DAS with single level optimization on the blur data formation. Right: Both methods with their own BOHB hyperparameters on the blur data formation.
    \label{fig:darts_single_level}}

\end{figure}
Several works, such as \cite{zela_understanding_2020}, investigate the instability of the bi-level approximation of DAS w.r.t. the weight initialization; the random initialization of the network weights can cause promising operations having poor initialization and thus tend to be entirely discarded during the architecture search.
We evaluate the impact of this initialization by modifying the DAS search, such that it only has to search for the optimal architecture parameters to build the resulting architecture.

For this \textit{DAS-single} approach we pre-train the operations $\{\textrm{learnable~gradient~descent}\}$ and $\{$~2-layer-CNN~$\}$ as baseline architectures consisting only of each operation respectively and keep the weights fixed. This is generally only possible for the feed-forward architectures that we consider and requires only a weak specialization between layers. Thereby, we avoid the random initialization of the operations weights in the DAS search. Figure \ref{fig:darts_single_level} shows the results of DAS-single search with BOHB-optimized hyperparameters for all operations. Notably, BOHB-optimized hyperparameters for the DAS-single one-shot validation (Fig. \ref{fig:darts_single_level} top) lead to a positive impact on the correlation of the one-shot and architecture validation PSNR using DAS-single and to a negative impact for DAS. In addition, when comparing DAS and DAS-single with their hyperparameters being individually optimized with BOHB with respect to their one-shot validation (Fig. \ref{fig:darts_single_level} bottom), DAS finds a higher architecture validation PSNR than DAS-single, whereas DAS-single becomes more robust against possible outliers, making this search less sensitive.

\subsection{Non-sequential Search Space}\label{exp:ablation}
In this section we additionally investigate the the stability of our DAS framework with respect to  hyperparameters within the non-sequential search space from \autoref{sec:exp_non_sequential} for the \textit{blur} data formation. We also visualize architectures found by our DAS approach for two different operation sets (Figure \ref{fig:found_wide_archs}).

\begin{figure}[h]
    \centering
    \includegraphics[width=0.48\textwidth]{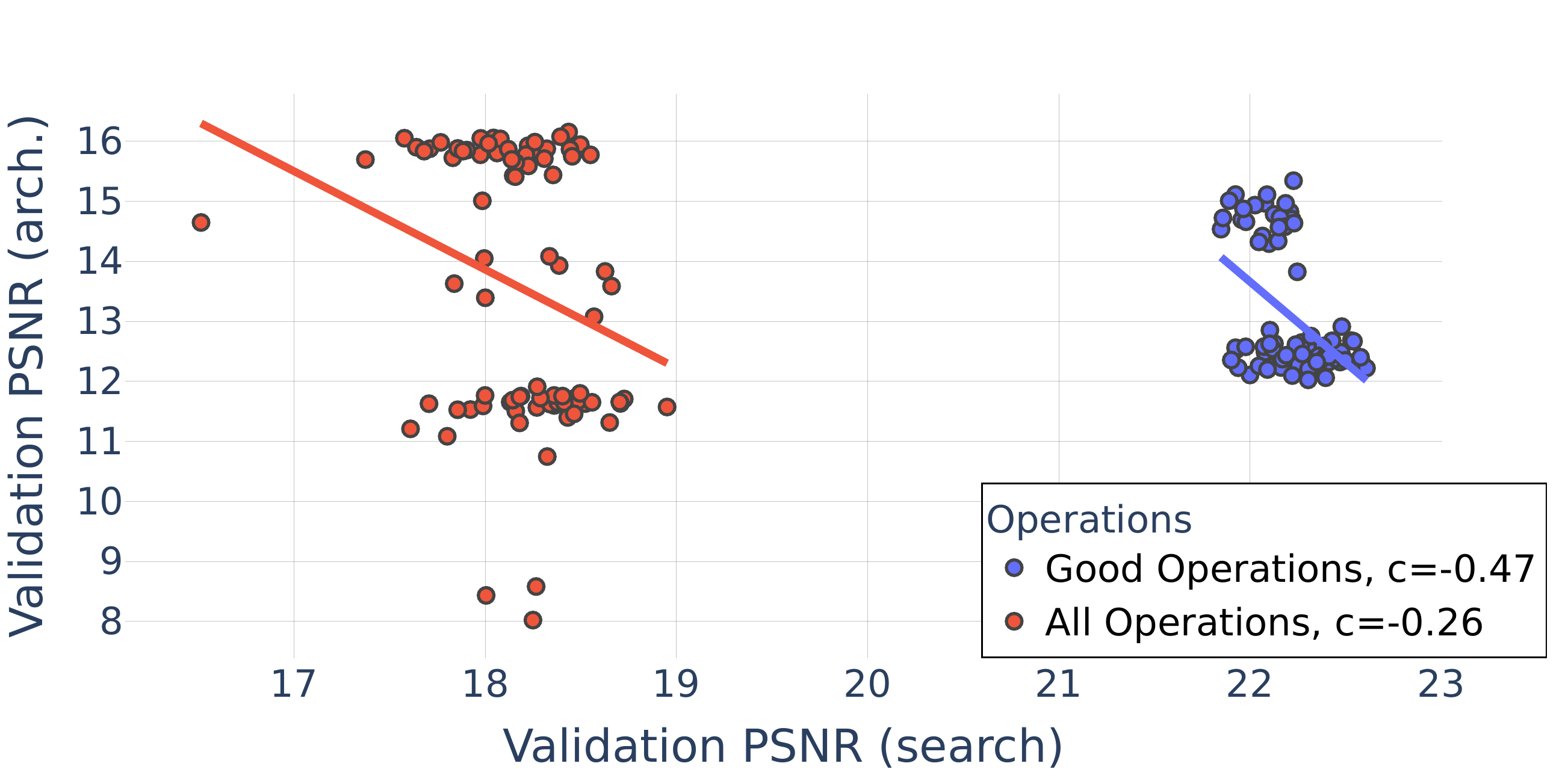}
    \includegraphics[width=0.48\textwidth]{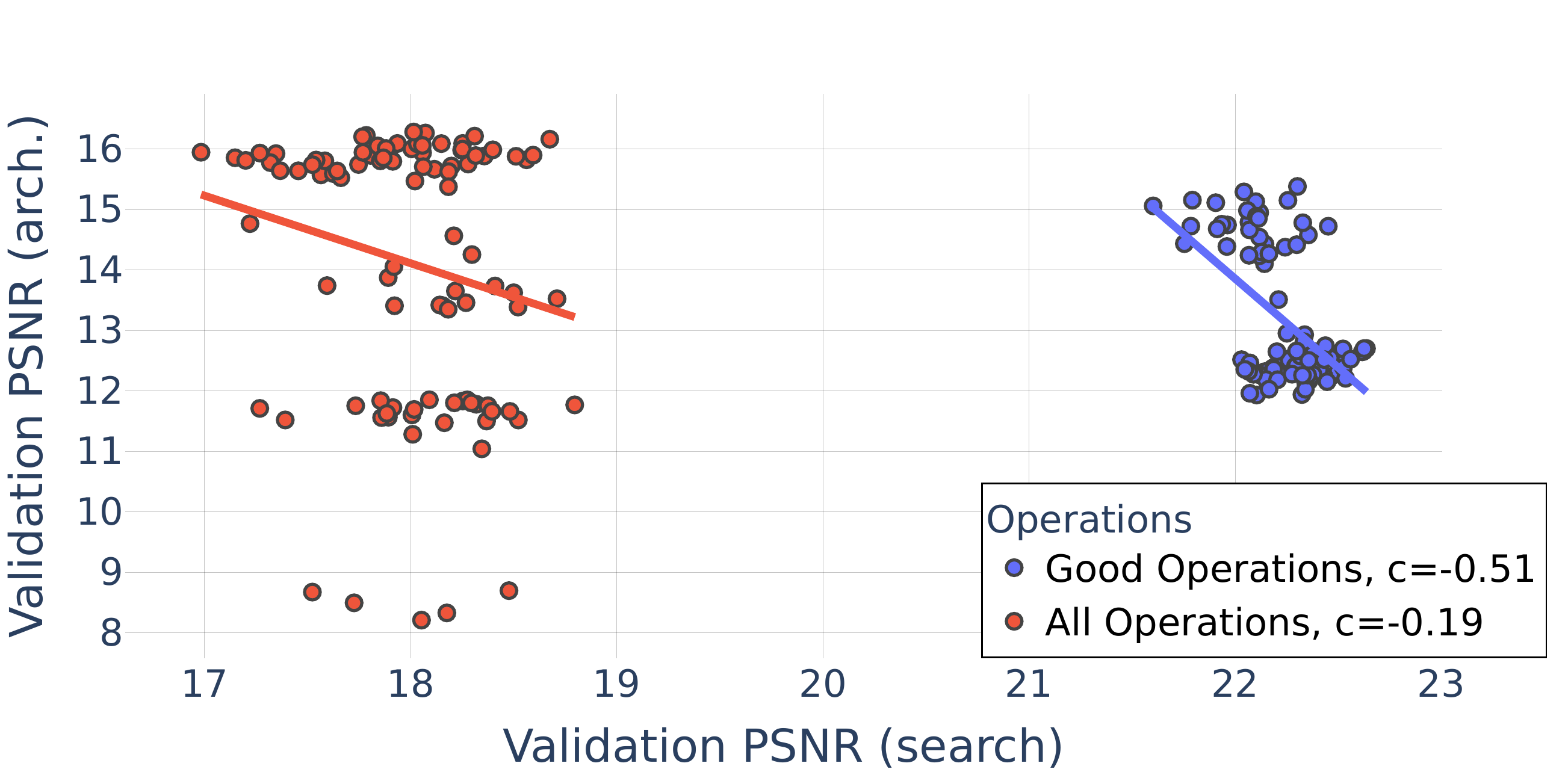}
    \caption{Scatter plot for the non-sequential DAS search space corresponding to \autoref{tab:widetabular_results_ablation}, with hyperparameters \textit{H1}(left) and \textit{H2} (right), showing architecture PSNR (y-axis) plotted against 1-shot validation PSNR (i.e. the validation performance on the DAS objective).
    \label{fig:correlated_wide_H}}
\end{figure}

\begin{figure}[h]
    \centering
    \includegraphics[width=0.48\textwidth]{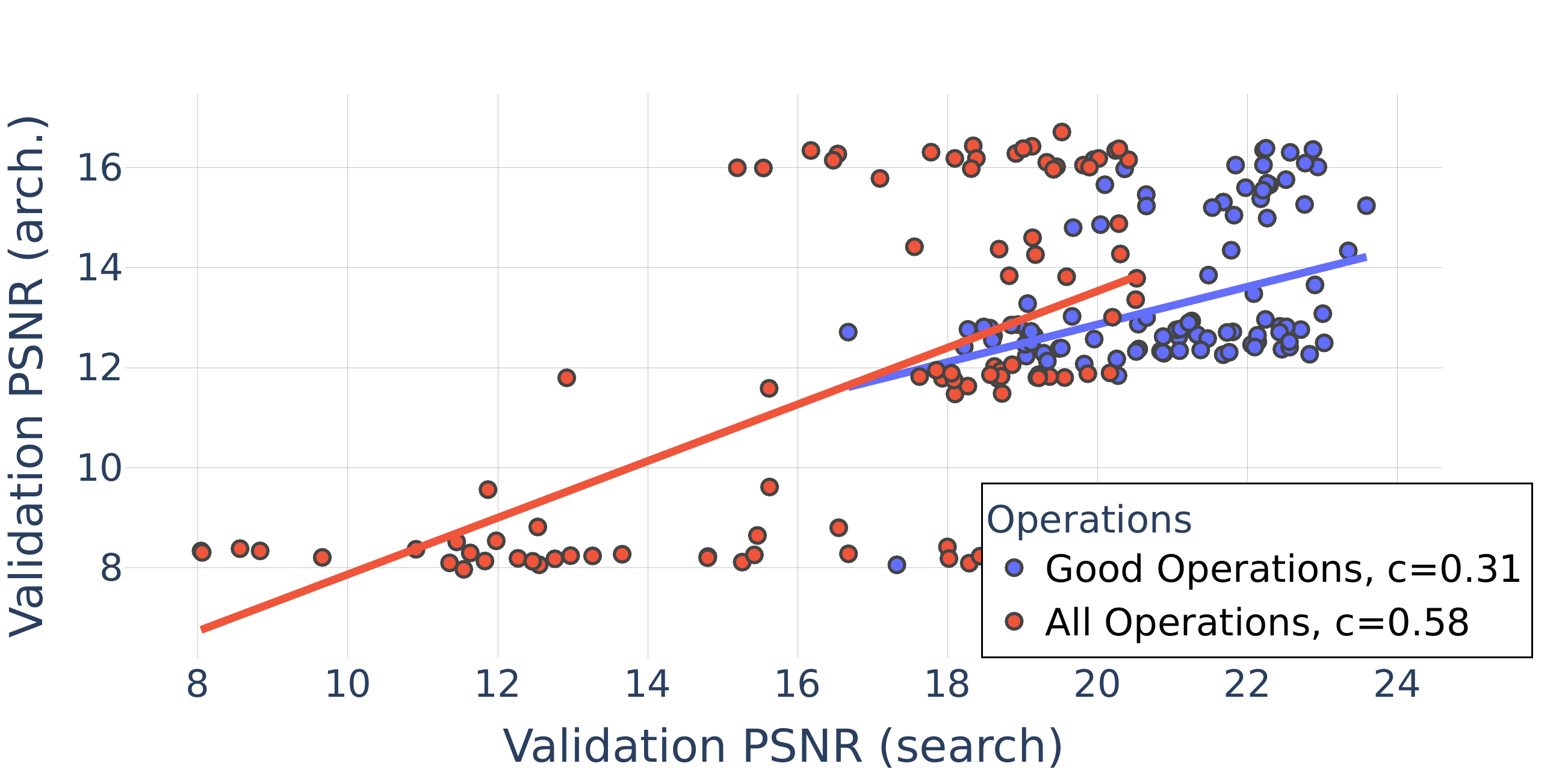}
    \includegraphics[width=0.48\textwidth]{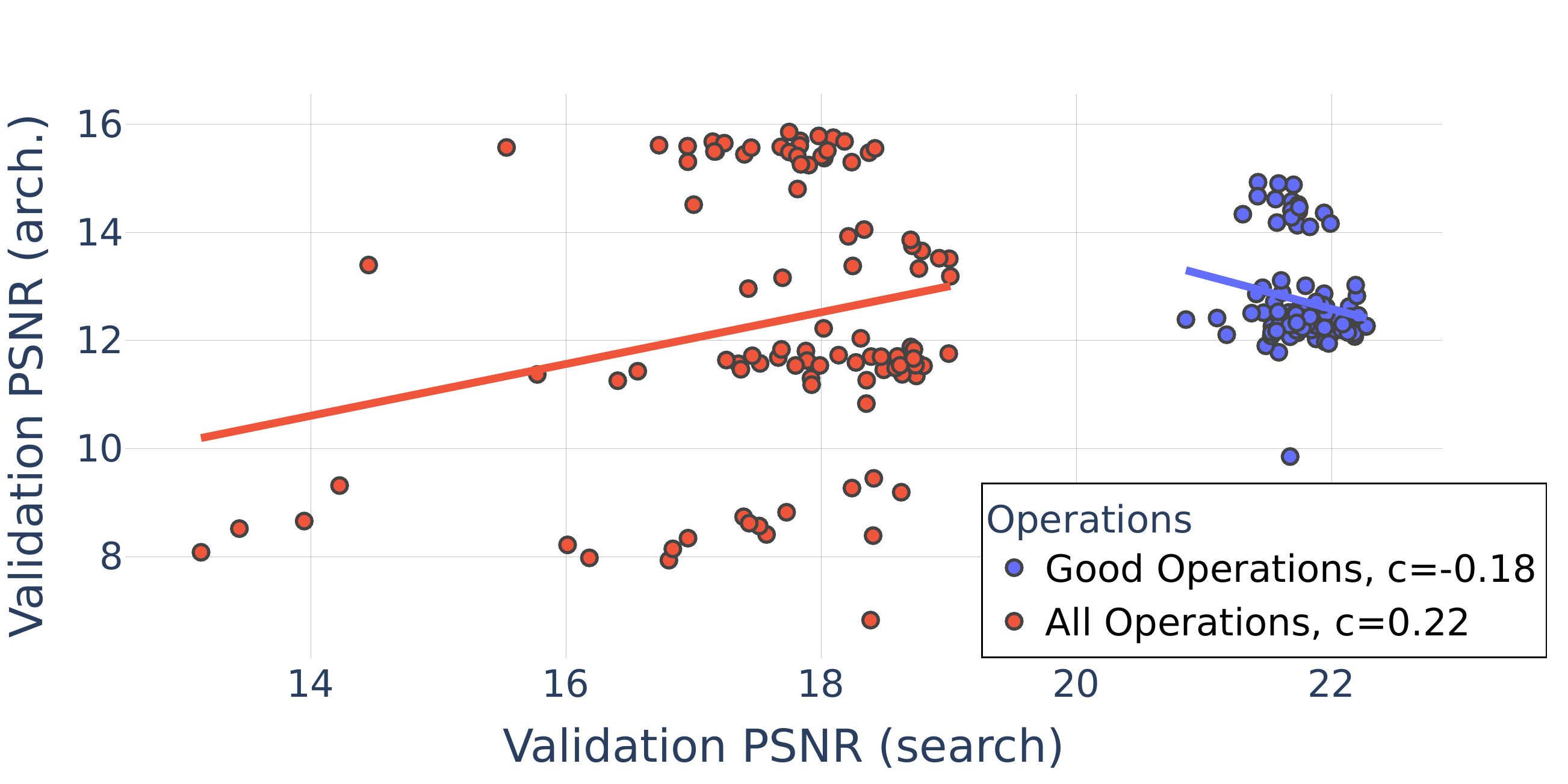}
    \includegraphics[width=0.48\textwidth]{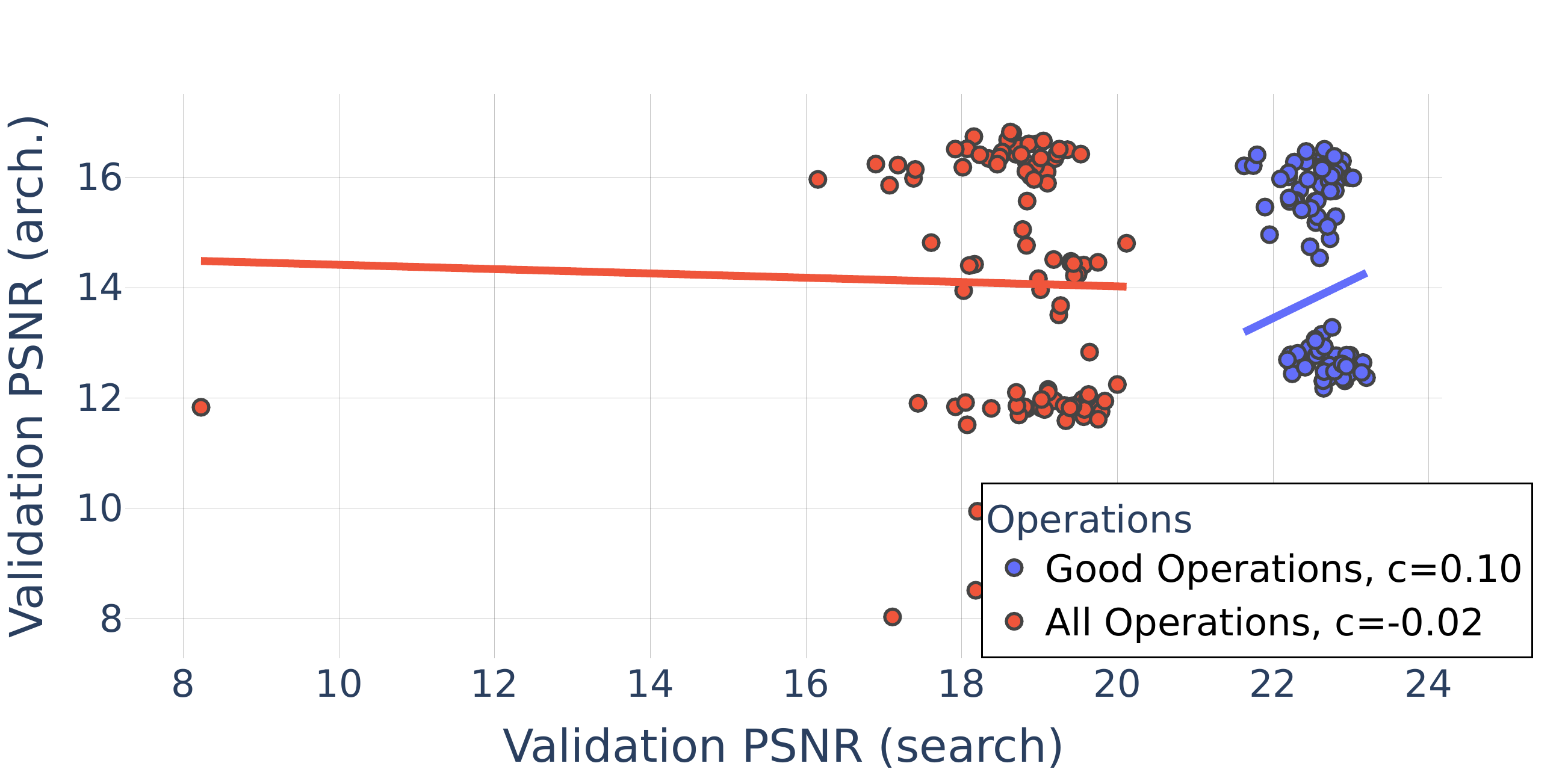}
    \caption{Scatter plot for the non-sequential DAS search space on blur with BOHB-optimized hyperparameters for this search space, showing architecture PSNR (y-axis) plotted against one-shot validation PSNR (x-axis). Top (Left): Blur with hyperparameters  \textit{BOHB-one-shot-Blur}. Top (Right): Blur with hyperparameters \textit{BOHB-one-shot-DS}. Bottom: Blur with hyperparameters \textit{BOHB-Blur}
    \label{fig:correlated_wide_Hbohb}}
\end{figure}

\begin{figure}[h]
    \centering
    \includegraphics[width=0.48\columnwidth]{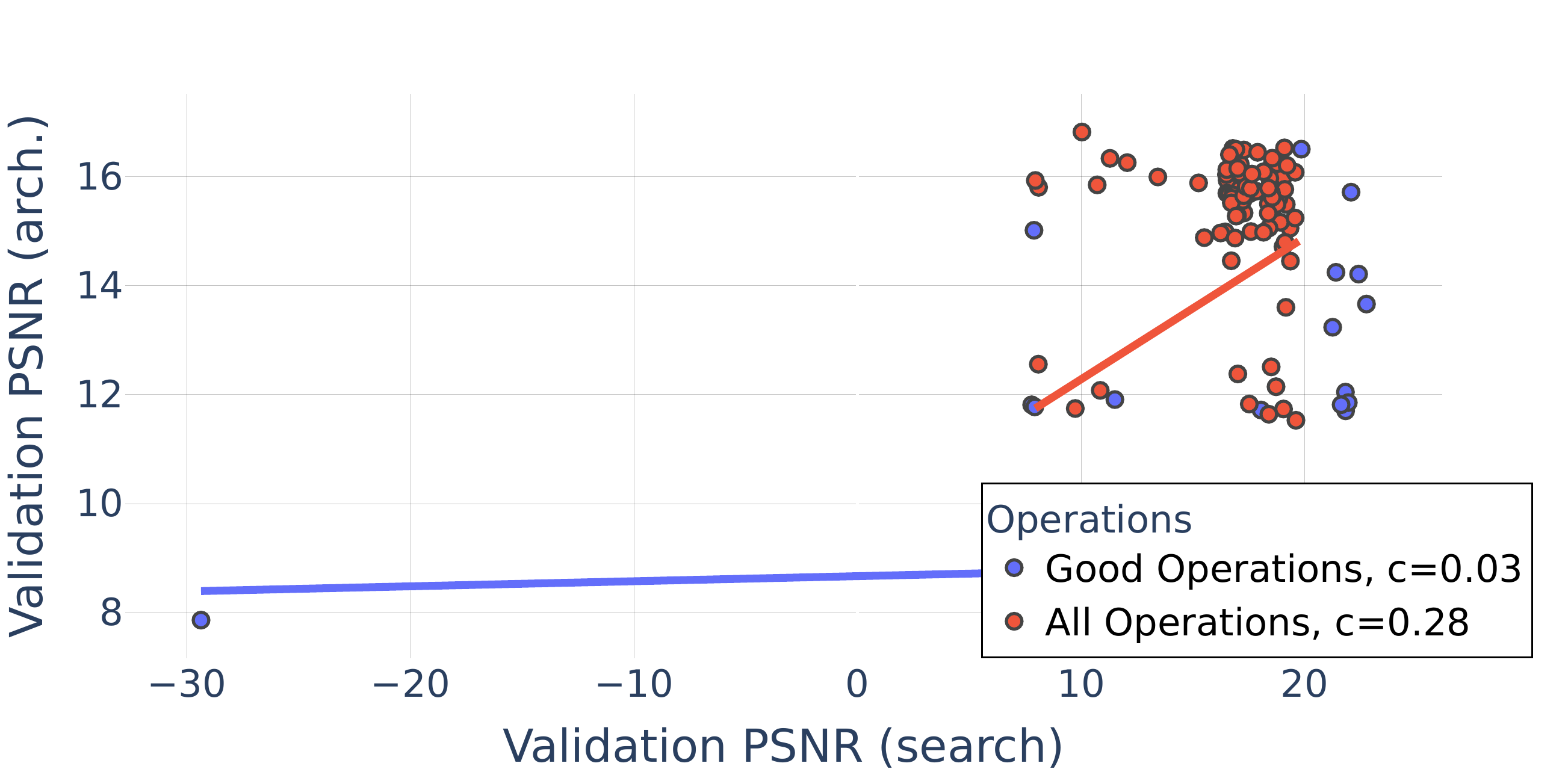}
    \includegraphics[width=0.48\columnwidth]{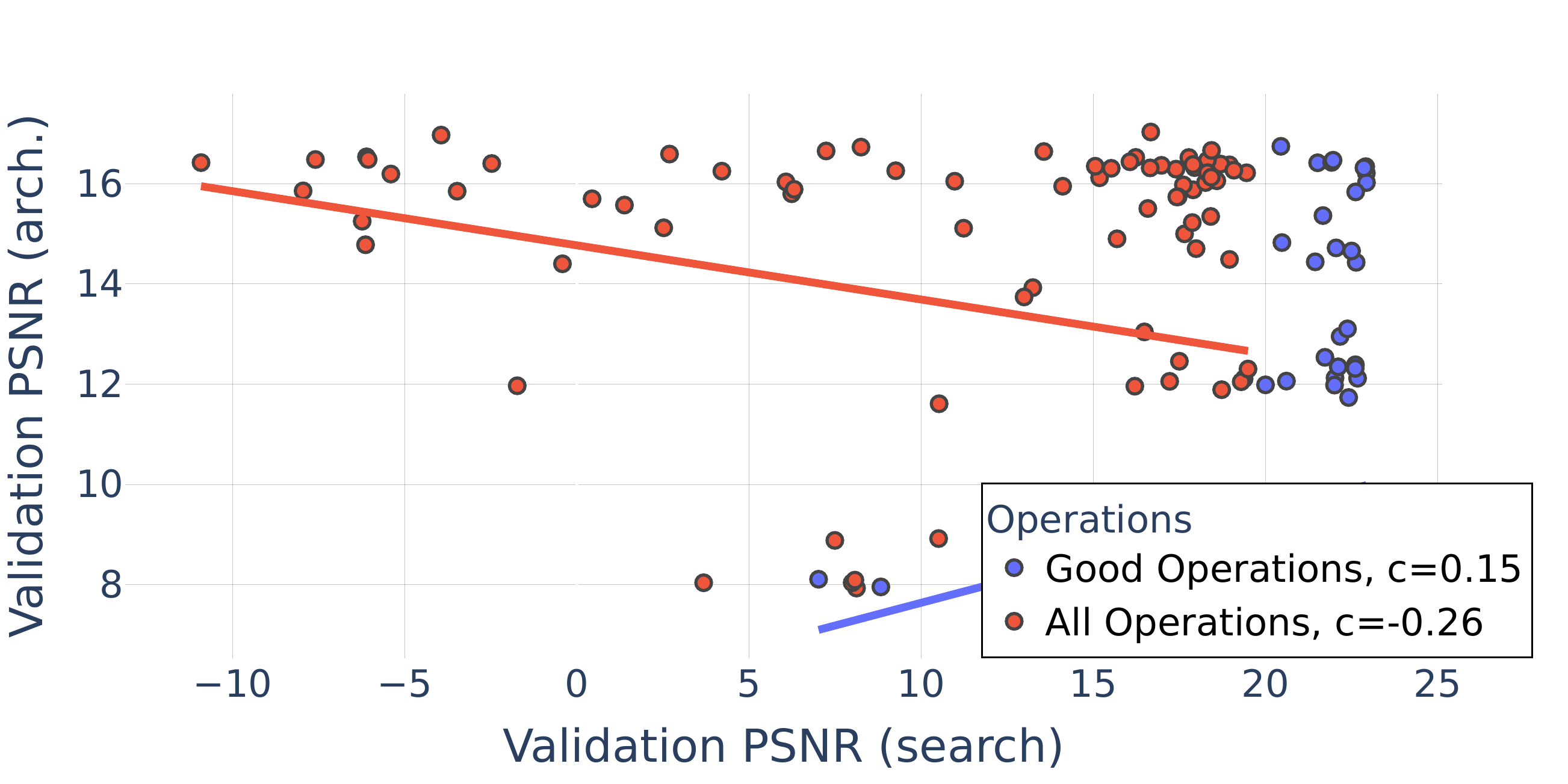}
    \caption{Scatter plot for the non-sequential DAS search space on blur with hyperparameters searched for this search space, showing architecture PSNR (y-axis) plotted against one-shot validation PSNR (x-axis). Left:  Blur with hyperparameters  \textit{BOHB-Non-Seq-one-shot-Blur} Right: Blur with hyperparameters searched for the final architecture performance \textit{BOHB-Non-Seq-Blur}.
    \label{fig:correlated_wide_wide}}
\end{figure}

\begin{table*}
    \centering
    \caption{DARTS-like \cite{liu_darts_2019} wide architecture validation PSNR values found in the 1D inverse problems setting with cosine data. Shown is the maximal, mean and median PSNR over 75 trials.
    \label{tab:widetabular_results_ablation}}
     \begin{tabular}{c@{\hspace{0.1cm}}|@{\hspace{0.1cm}}c@{\hspace{0.1cm}}|@{\hspace{0.3cm}}c@{\hspace{0.3cm}}c@{\hspace{0.3cm}}c@{\hspace{0.3cm}}|@{\hspace{0.3cm}}c@{\hspace{0.3cm}}c@{\hspace{0.3cm}}c}
    \toprule
        Data & Hyperparameters & \multicolumn{6}{c}{Architecture Validation (PSNR)}\\
        & & \multicolumn{3}{c@{\hspace{0.3cm}}|@{\hspace{0.3cm}}}{Good Ops.} & \multicolumn{3}{c}{All} \\
        & & Max. & Mean & Med. & Max. & Mean & Med.\\
        \midrule
        \multirow{7}{*}{Blur}& H1 &15.34 & 13.08  &   12.51& 16.15 & 13.56 & 13.73 \\
        & H2 & 15.38 & 13.17  & 12.52  & 16.28 & 14.11 & 15.58 \\
         & BOHB-one-shot-Blur & 16.38 &    13.25&   12.76& 16.71 & 11.73 & 11.8 \\
         & BOHB-one-shot-DS &14.93 &  12.73 & 12.44  & 15.86 & 12.37 & 11.72 \\
         & BOHB-Blur & 16.5 & 13.83  &  13.06 & 16.82 & 14.07 & 14.44 \\
         & BOHB-Non-Seq-one-shot-Blur & 16.50 & 8.84 &  8.09 & 16.82 & 13.96 & 15.5 \\
         & BOHB-Non-Seq-Blur & 16.74 & 9.73 &  8.11 & 17.03 & 13.45 & 15.42 \\
        \bottomrule
    \end{tabular}
\end{table*}

To investigate the hyperparameter stability further for this non-sequential search space, we conduct experiments using the same BOHB-optimized hyperparameters as in \autoref{sec:correlation} and additionally included BOHB-optimized hyperparameters for this non-sequential search space for first targeting the one-shot validation performance (BOHB-Non-Seq-one-shot-Blur) and second targeting the final architecture performance (BOHB-Non-Seq-Blur). 
\autoref{tab:widetabular_results_ablation} however shows similar results as in \autoref{sec:correlation}: changing the hyperparameters in this non-sequential search space does not improve the stability of the search process.
\autoref{fig:correlated_wide_H} shows all trials for the non-sequential search space for the manually chosen hyperparameters \textit{H1} and \textit{H2}. This plot clarifies further, that the search space change does not improve the DAS search process. The correlation between the one-shot validation and the architecture validation even becomes negative. Yet, these plots also show 2 different ``failure'' cases for both operations sets, only beneficial operations and all operations, and both data formations: The validation PSNR is stable, whereas the architecture validation performance is clustered in two different regions, one being very low and the other being around 15 PSNR. Note, the mean architecture validation PSNR for all operations in the sequential search space from \autoref{sec:correlation} in \autoref{tab:tabular_results} is also around 15 PSNR.

For additional visualization, we also display the results using BOHB found hyperparameters in the sequential search space in \autoref{fig:correlated_wide_Hbohb} as well as BOHB found hyperparameters tuned for this non-sequential search space in \autoref{fig:correlated_wide_wide}. However, hyperparameter search for the non-sequential search space via BOHB on both the one-shot validation performance and the architecture performance as a target, does not actually improve the stability of the search for this new search space, as demonstrated in \autoref{fig:correlated_wide_wide}. Accordingly we find on the one hand that the findings in the previous section \ref{sec:results} regarding non-applicability of DAS as a one-shot model for inverse problems translate to a cell-based search space and on the other hand (investigating the overall performance metrics for both search spaces), that the sequential search space appears to be a helpful prior for architecture search for inverse problems, given that its PSNR scores are overall higher.

\section{Conclusions}
In this paper we analyzed DAS in a systematic study on one-dimensional inverse reconstruction problems. In this setting, we show that DAS improves over a random search baseline by a significant margin, especially if the available set of beneficial operation is not determined in advance. In our analysis, we make the following findings: While it is possible to find well-performing architectures using DAS, multiple runs of the same setting yield a high variance, challenging the common understanding of DAS as a one-shot method. Moreover, the ability to find well-performing architectures is highly dependent on the specific choice of hyperparameters. Unfortunately, judging the success of any DAS-based model right after the one-shot training is difficult, since a strong correlation to the actual architecture performance is missing. As such, even automatic hyperparameter searches such as BOHB cannot faithfully be applied to the one-shot loss.

Therefore, we emphasize for the future the necessity to (1) 
    look at a full statistical evaluation of DAS performances over multiple trials, in all applications where this is feasible, and 
     (2) show a reasonable correlation between the search and final architecture performances for any method that reports improved results based on a more faithful minimization of the one-shot DAS objective.

\section*{Acknowledgments}
JL and MK acknowledge support by the German Federal Ministry of Education and Research Foundation via the project DeToL.

\bibliographystyle{plainnat}
\bibliography{zotero_library,manual_references}

\begin{thebibliography}{35}
\providecommand{\natexlab}[1]{#1}
\providecommand{\url}[1]{\texttt{#1}}
\expandafter\ifx\csname urlstyle\endcsname\relax
  \providecommand{\doi}[1]{doi: #1}\else
  \providecommand{\doi}{doi: \begingroup \urlstyle{rm}\Url}\fi

\bibitem[Adler and {\"O}ktem(2018)]{adler_learned_2018}
Jonas Adler and Ozan {\"O}ktem.
\newblock Learned {{Primal-Dual Reconstruction}}.
\newblock \emph{IEEE Transactions on Medical Imaging}, 37\penalty0
  (6):\penalty0 1322--1332, June 2018.
\newblock ISSN 1558-254X.
\newblock \doi{10.1109/TMI.2018.2799231}.

\bibitem[Aggarwal et~al.(2019)Aggarwal, Mani, and Jacob]{aggarwal_modl_2019}
Hemant~K. Aggarwal, Merry~P. Mani, and Mathews Jacob.
\newblock {{MoDL}}: {{Model-Based Deep Learning Architecture}} for {{Inverse
  Problems}}.
\newblock \emph{IEEE Transactions on Medical Imaging}, 38\penalty0
  (2):\penalty0 394--405, February 2019.
\newblock ISSN 1558-254X.
\newblock \doi{10.1109/TMI.2018.2865356}.

\bibitem[Akimoto et~al.(2019)Akimoto, Shirakawa, Yoshinari, Uchida, Saito, and
  Nishida]{Akimoto_2019_Adaptive}
Youhei Akimoto, Shinichi Shirakawa, Nozomu Yoshinari, Kento Uchida, Shota
  Saito, and Kouhei Nishida.
\newblock Adaptive stochastic natural gradient method for one-shot neural
  architecture search.
\newblock In \emph{Proceedings of the 36th International Conference on Machine
  Learning, {ICML} 2019, 9-15 June 2019, Long Beach, California, {USA}},
  volume~97 of \emph{Proceedings of Machine Learning Research}, pages 171--180.
  {PMLR}, 2019.
\newblock URL \url{http://proceedings.mlr.press/v97/akimoto19a.html}.

\bibitem[Cai et~al.(2019)Cai, Zhu, and Han]{cai_proxylessnas_2019}
Han Cai, Ligeng Zhu, and Song Han.
\newblock {{ProxylessNAS}}: {{Direct Neural Architecture Search}} on {{Target
  Task}} and {{Hardware}}.
\newblock \emph{arXiv:1812.00332 [cs, stat]}, February 2019.
\newblock URL \url{http://arxiv.org/abs/1812.00332}.

\bibitem[Chen and Hsieh(2020)]{Chen_2020_SDARTS}
Xiangning Chen and Cho{-}Jui Hsieh.
\newblock Stabilizing differentiable architecture search via perturbation-based
  regularization.
\newblock In \emph{Proceedings of the 37th International Conference on Machine
  Learning, {ICML} 2020, 13-18 July 2020, Virtual Event}, volume 119 of
  \emph{Proceedings of Machine Learning Research}, pages 1554--1565. {PMLR},
  2020.
\newblock URL \url{http://proceedings.mlr.press/v119/chen20f.html}.

\bibitem[Chen et~al.(2019)Chen, Xie, Wu, and Tian]{Chen_2019_PDARTS}
Xin Chen, Lingxi Xie, Jun Wu, and Qi~Tian.
\newblock Progressive differentiable architecture search: Bridging the depth
  gap between search and evaluation.
\newblock In \emph{2019 {IEEE/CVF} International Conference on Computer Vision,
  {ICCV} 2019, Seoul, Korea (South), October 27 - November 2, 2019}, pages
  1294--1303. {IEEE}, 2019.
\newblock \doi{10.1109/ICCV.2019.00138}.
\newblock URL \url{https://doi.org/10.1109/ICCV.2019.00138}.

\bibitem[Chu et~al.(2020)Chu, Zhou, Zhang, and Li]{Chu_2020_FairDarts}
Xiangxiang Chu, Tianbao Zhou, Bo~Zhang, and Jixiang Li.
\newblock Fair {DARTS:} eliminating unfair advantages in differentiable
  architecture search.
\newblock In Andrea Vedaldi, Horst Bischof, Thomas Brox, and Jan{-}Michael
  Frahm, editors, \emph{Computer Vision - {ECCV} 2020 - 16th European
  Conference, Glasgow, UK, August 23-28, 2020, Proceedings, Part {XV}}, volume
  12360 of \emph{Lecture Notes in Computer Science}, pages 465--480. Springer,
  2020.
\newblock \doi{10.1007/978-3-030-58555-6\_28}.
\newblock URL \url{https://doi.org/10.1007/978-3-030-58555-6\_28}.

\bibitem[Dong and Yang(2019)]{dong_searching_2019}
Xuanyi Dong and Yi~Yang.
\newblock Searching for {{A Robust Neural Architecture}} in {{Four GPU Hours}}.
\newblock \emph{arXiv:1910.04465 [cs]}, October 2019.
\newblock URL \url{http://arxiv.org/abs/1910.04465}.

\bibitem[Falkner et~al.(2018)Falkner, Klein, and Hutter]{falkner_bohb_2018}
Stefan Falkner, Aaron Klein, and Frank Hutter.
\newblock {{BOHB}}: {{Robust}} and {{Efficient Hyperparameter Optimization}} at
  {{Scale}}.
\newblock In \emph{International {{Conference}} on {{Machine Learning}}}, pages
  1437--1446. {PMLR}, July 2018.
\newblock URL \url{http://proceedings.mlr.press/v80/falkner18a.html}.

\bibitem[Gregor and LeCun(2010)]{lista}
Karol Gregor and Yann LeCun.
\newblock Learning fast approximations of sparse coding.
\newblock In \emph{Proceedings of the 27th International Conference on
  International Conference on Machine Learning}, page 399–406, 2010.

\bibitem[Hammernik et~al.(2017)Hammernik, W{\"u}rfl, Pock, and
  Maier]{hammernik_deep_2017}
Kerstin Hammernik, Tobias W{\"u}rfl, Thomas Pock, and Andreas Maier.
\newblock {A Deep Learning Architecture for Limited-Angle Computed Tomography
  Reconstruction}.
\newblock In \emph{{Bildverarbeitung f\"ur die Medizin 2017}}, {Informatik
  aktuell}, pages 92--97. {Springer Berlin Heidelberg}, 2017.
\newblock ISBN 978-3-662-54345-0.

\bibitem[Hammernik et~al.(2018)Hammernik, Klatzer, Kobler, Recht, Sodickson,
  Pock, and Knoll]{hammernik_learning_2018}
Kerstin Hammernik, Teresa Klatzer, Erich Kobler, Michael~P. Recht, Daniel~K.
  Sodickson, Thomas Pock, and Florian Knoll.
\newblock Learning a variational network for reconstruction of accelerated
  {{MRI}} data.
\newblock \emph{Magnetic Resonance in Medicine}, 79\penalty0 (6):\penalty0
  3055--3071, June 2018.
\newblock ISSN 1522-2594.
\newblock \doi{10.1002/mrm.26977}.

\bibitem[He et~al.(2020)He, Ye, Shen, and Zhang]{He_2020_Milenas}
Chaoyang He, Haishan Ye, Li~Shen, and Tong Zhang.
\newblock Milenas: Efficient neural architecture search via mixed-level
  reformulation.
\newblock In \emph{2020 {IEEE/CVF} Conference on Computer Vision and Pattern
  Recognition, {CVPR} 2020, Seattle, WA, USA, June 13-19, 2020}, pages
  11990--11999. {IEEE}, 2020.
\newblock \doi{10.1109/CVPR42600.2020.01201}.
\newblock URL \url{https://doi.org/10.1109/CVPR42600.2020.01201}.

\bibitem[He et~al.(2015)He, Zhang, Ren, and Sun]{he_deep_2015}
Kaiming He, Xiangyu Zhang, Shaoqing Ren, and Jian Sun.
\newblock Deep {{Residual Learning}} for {{Image Recognition}}.
\newblock \emph{arXiv:1512.03385 [cs]}, December 2015.
\newblock URL \url{http://arxiv.org/abs/1512.03385}.

\bibitem[Kandasamy et~al.(2018)Kandasamy, Neiswanger, Schneider, P{\'{o}}czos,
  and Xing]{Kandasamy2018}
Kirthevasan Kandasamy, Willie Neiswanger, Jeff Schneider, Barnab{\'{a}}s
  P{\'{o}}czos, and Eric~P. Xing.
\newblock Neural architecture search with bayesian optimisation and optimal
  transport.
\newblock In \emph{NeurIPS}, 2018.

\bibitem[Klatzer et~al.(2016)Klatzer, Hammernik, Knobelreiter, and
  Pock]{klatzer_learning_2016}
Teresa Klatzer, Kerstin Hammernik, Patrick Knobelreiter, and Thomas Pock.
\newblock Learning joint demosaicing and denoising based on sequential energy
  minimization.
\newblock In \emph{2016 {{IEEE International Conference}} on {{Computational
  Photography}} ({{ICCP}})}, pages 1--11, May 2016.
\newblock \doi{10.1109/ICCPHOT.2016.7492871}.

\bibitem[Kobler et~al.(2017)Kobler, Klatzer, Hammernik, and
  Pock]{VariationalNetworks}
Erich Kobler, Teresa Klatzer, Kerstin Hammernik, and Thomas Pock.
\newblock Variational networks: Connecting variational methods and deep
  learning.
\newblock In \emph{Pattern Recognition}, Lecture Notes in Computer Science,
  pages 281--293. Springer, 2017.
\newblock ISBN 978-3-319-66708-9.
\newblock \doi{10.1007/978-3-319-66709-6_23}.

\bibitem[Li et~al.(2018)Li, Jamieson, DeSalvo, Rostamizadeh, and
  Talwalkar]{li_hyperband_2018}
Lisha Li, Kevin Jamieson, Giulia DeSalvo, Afshin Rostamizadeh, and Ameet
  Talwalkar.
\newblock Hyperband: {{A Novel Bandit-Based Approach}} to {{Hyperparameter
  Optimization}}.
\newblock \emph{Journal of Machine Learning Research}, 18\penalty0
  (185):\penalty0 1--52, 2018.
\newblock ISSN 1533-7928.
\newblock URL \url{http://jmlr.org/papers/v18/16-558.html}.

\bibitem[Liu et~al.(2018)Liu, Zoph, Neumann, Shlens, Hua, Li, Fei{-}Fei,
  Yuille, Huang, and Murphy]{Liu_2018_progressivenas}
Chenxi Liu, Barret Zoph, Maxim Neumann, Jonathon Shlens, Wei Hua, Li{-}Jia Li,
  Li~Fei{-}Fei, Alan~L. Yuille, Jonathan Huang, and Kevin Murphy.
\newblock Progressive neural architecture search.
\newblock In \emph{Computer Vision - {ECCV} 2018 - 15th European Conference,
  Munich, Germany, September 8-14, 2018, Proceedings, Part {I}}, volume 11205
  of \emph{Lecture Notes in Computer Science}, pages 19--35. Springer, 2018.
\newblock \doi{10.1007/978-3-030-01246-5\_2}.
\newblock URL \url{https://doi.org/10.1007/978-3-030-01246-5\_2}.

\bibitem[Liu et~al.(2019)Liu, Simonyan, and Yang]{liu_darts_2019}
Hanxiao Liu, Karen Simonyan, and Yiming Yang.
\newblock {{DARTS}}: {{Differentiable Architecture Search}}.
\newblock \emph{arXiv:1806.09055 [cs, stat]}, April 2019.
\newblock URL \url{http://arxiv.org/abs/1806.09055}.

\bibitem[Lukasik et~al.(2021)Lukasik, Friede, Zela, Hutter, and
  Keuper]{Lukasik_2021_svge}
Jovita Lukasik, David Friede, Arber Zela, Frank Hutter, and Margret Keuper.
\newblock Smooth variational graph embeddings for efficient neural architecture
  search, 2021.

\bibitem[Pham et~al.(2018)Pham, Guan, Zoph, Le, and Dean]{Pham_2018}
Hieu Pham, Melody~Y. Guan, Barret Zoph, Quoc~V. Le, and Jeff Dean.
\newblock Efficient neural architecture search via parameter sharing.
\newblock In Jennifer~G. Dy and Andreas Krause, editors, \emph{Proceedings of
  the 35th International Conference on Machine Learning, {ICML} 2018,
  Stockholmsm{\"{a}}ssan, Stockholm, Sweden, July 10-15, 2018}, volume~80 of
  \emph{Proceedings of Machine Learning Research}, pages 4092--4101. {PMLR},
  2018.

\bibitem[Riegler et~al.(2016)Riegler, R{\"u}ther, and
  Bischof]{riegler_atgv-net:_2016}
Gernot Riegler, Matthias R{\"u}ther, and Horst Bischof.
\newblock Atgv-net: {{Accurate}} depth super-resolution.
\newblock In \emph{European {{Conference}} on {{Computer Vision}}}, pages
  268--284. {Springer}, 2016.
\newblock URL
  \url{http://link.springer.com/chapter/10.1007/978-3-319-46487-9_17}.

\bibitem[Ru et~al.(2020)Ru, Wan, Dong, and Osborne]{Ru2020NeuralAS}
Binxin Ru, Xingchen Wan, Xiaowen Dong, and Michael Osborne.
\newblock Neural architecture search using bayesian optimisation with
  weisfeiler-lehman kernel.
\newblock \emph{ArXiv}, abs/2006.07556, 2020.

\bibitem[Schmidt and Roth(2014)]{shrinkageFields}
Uwe Schmidt and Stefan Roth.
\newblock Shrinkage fields for effective image restoration.
\newblock In \emph{IEEE Conference on Computer Vision and Pattern Recognition},
  pages 2774--2781, 2014.
\newblock \doi{10.1109/CVPR.2014.349}.

\bibitem[White et~al.(2019)White, Neiswanger, and Savani]{white2019bananas}
Colin White, Willie Neiswanger, and Yash Savani.
\newblock Bananas: Bayesian optimization with neural architectures for neural
  architecture search.
\newblock \emph{arXiv preprint arXiv:1910.11858}, 2019.

\bibitem[Wu et~al.(2021)Wu, Liu, Huang, Zhang, and
  Gool]{Wu_2021_SparseSupernet}
Yan Wu, Aoming Liu, Zhiwu Huang, Siwei Zhang, and Luc~Van Gool.
\newblock Neural architecture search as sparse supernet.
\newblock In \emph{Thirty-Fifth {AAAI} Conference on Artificial Intelligence,
  {AAAI} 2021, Thirty-Third Conference on Innovative Applications of Artificial
  Intelligence, {IAAI} 2021, The Eleventh Symposium on Educational Advances in
  Artificial Intelligence, {EAAI} 2021, Virtual Event, February 2-9, 2021},
  pages 10379--10387. {AAAI} Press, 2021.
\newblock URL \url{https://ojs.aaai.org/index.php/AAAI/article/view/17243}.

\bibitem[Xie et~al.(2019)Xie, Zheng, Liu, and Lin]{Xie_2019_SNAS}
Sirui Xie, Hehui Zheng, Chunxiao Liu, and Liang Lin.
\newblock {SNAS:} stochastic neural architecture search.
\newblock In \emph{7th International Conference on Learning Representations,
  {ICLR} 2019, New Orleans, LA, USA, May 6-9, 2019}. OpenReview.net, 2019.
\newblock URL \url{https://openreview.net/forum?id=rylqooRqK7}.

\bibitem[Xu et~al.(2020)Xu, Xie, Zhang, Chen, Qi, Tian, and
  Xiong]{XU_2020_PCdarts}
Yuhui Xu, Lingxi Xie, Xiaopeng Zhang, Xin Chen, Guo{-}Jun Qi, Qi~Tian, and
  Hongkai Xiong.
\newblock {PC-DARTS:} partial channel connections for memory-efficient
  architecture search.
\newblock In \emph{8th International Conference on Learning Representations,
  {ICLR} 2020, Addis Ababa, Ethiopia, April 26-30, 2020}. OpenReview.net, 2020.
\newblock URL \url{https://openreview.net/forum?id=BJlS634tPr}.

\bibitem[Yang et~al.(2020)Yang, Esperan{\c c}a, and Carlucci]{yang_nas_2020}
Antoine Yang, Pedro~M. Esperan{\c c}a, and Fabio~M. Carlucci.
\newblock {{NAS}} evaluation is frustratingly hard.
\newblock In \emph{Eighth {{International Conference}} on {{Learning
  Representations}}}, April 2020.
\newblock URL \url{https://iclr.cc/virtual_2020/poster_HygrdpVKvr.html}.

\bibitem[Zela et~al.(2020)Zela, Elsken, Saikia, Marrakchi, Brox, and
  Hutter]{zela_understanding_2020}
Arber Zela, Thomas Elsken, Tonmoy Saikia, Yassine Marrakchi, Thomas Brox, and
  Frank Hutter.
\newblock Understanding and {{Robustifying Differentiable Architecture
  Search}}.
\newblock \emph{arXiv:1909.09656 [cs, stat]}, January 2020.
\newblock URL \url{http://arxiv.org/abs/1909.09656}.

\bibitem[Zhang et~al.(2017)Zhang, Zuo, Chen, Meng, and
  Zhang]{zhang_beyond_2017}
Kai Zhang, Wangmeng Zuo, Yunjin Chen, Deyu Meng, and Lei Zhang.
\newblock Beyond a {{Gaussian Denoiser}}: {{Residual Learning}} of {{Deep CNN}}
  for {{Image Denoising}}.
\newblock \emph{IEEE Transactions on Image Processing}, 26\penalty0
  (7):\penalty0 3142--3155, July 2017.
\newblock ISSN 1057-7149.
\newblock \doi{10.1109/TIP.2017.2662206}.

\bibitem[Zhang et~al.(2021)Zhang, Su, Pan, Chang, Abbasnejad, and
  Haffari]{zhang_2021_idarts}
Miao Zhang, Steven Su, Shirui Pan, Xiaojun Chang, Ehsan Abbasnejad, and Reza
  Haffari.
\newblock idarts: Differentiable architecture search with stochastic implicit
  gradients, 2021.

\bibitem[Zoph and Le(2017)]{Zoph2017}
Barret Zoph and Quoc~V. Le.
\newblock Neural architecture search with reinforcement learning.
\newblock In \emph{ICLR}, 2017.

\bibitem[Zoph et~al.(2018)Zoph, Vasudevan, Shlens, and Le]{zoph2018learning}
Barret Zoph, Vijay Vasudevan, Jonathon Shlens, and Quoc~V Le.
\newblock Learning transferable architectures for scalable image recognition.
\newblock In \emph{CVPR}, 2018.

\end{thebibliography}

\appendix

\section*{Appendices}

\section{Visualizations} 
\begin{figure*}[h]
    \centering
    \includegraphics[height= 0.1\textwidth, width=\textwidth]{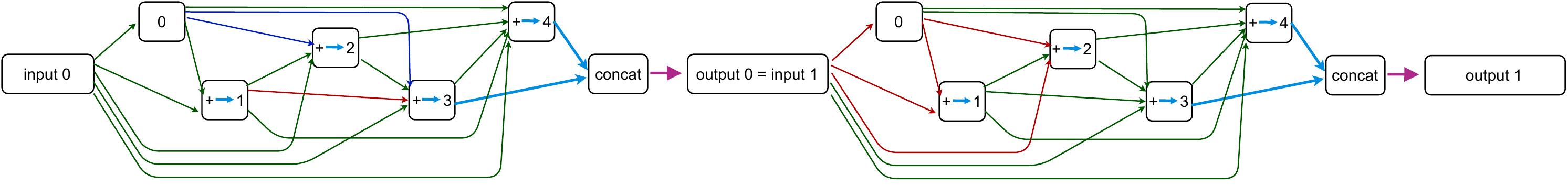} \\
    \includegraphics[height= 0.1\textwidth, width=\textwidth]{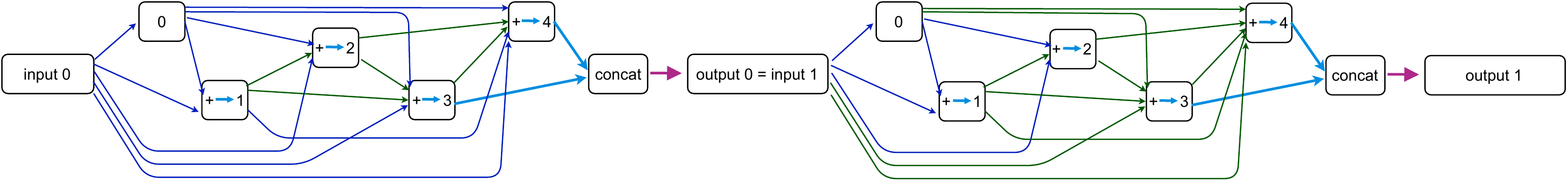}
    \caption{Found architectures in the non-sequential search space for two different operation sets for the data formation blur. Hyperparameter H1 is used for these searches. Top: all operations. Bottom: only beneficial operations.
    \label{fig:found_wide_archs}}
\end{figure*}
In this section we visualize in \autoref{fig:found_wide_archs}  two found architectures using the H1 hyperparameters for the operation sets ``all operations'' and ``only good operations'' for the data formation blur in the non-sequential search space from the experiments in \autoref{sec:results}.

\section{Hyperparameters} \label{supp:HP}
In this section we show the hyperparameters used for our experiments in the main paper. In \autoref{table:H1_H2} are the manually chosen hyperparameters H1 and H2 from \autoref{tab:tabular_results} 
in the main paper. In addition \autoref{table:HP_BOHB} lists all BOHB optimized hyperparameters for the data formations blur and downsampling as well as the hyperparameters optimized for the final architecture performance and also for the DAS-single method; the search range for the BOHB search is given in the second column. In \autoref{table:HP_BOHB_wide} the BOHB search hyperparameters for the non-sequential search space from \autoref{exp:ablation} are listed.
\autoref{table:GH} lists all other general hyperparameters used for our experiments.
\begin{table}[hb]
\centering
\caption{General Hyperparameters. \label{table:GH}}
\begin{tabular}{@{}lc@{}}
\toprule
 Hyperparameter       & Default Value \\ 
Epochs                    & $50$ \\
Batch size                & $128$ \\
Noise Level               & $0.10$\\
\bottomrule
\end{tabular}
\end{table}

\begin{table}
\centering
\caption{Manually chosen hyperparameters H1 and H2 \label{table:H1_H2}}

\small{
\begin{tabular}{@{}lcc@{}}
\toprule
Hyperparameter       &  H1 & H2\\ \midrule
Param. learning rate &  $0.001$      &  $0.001$       \\
Param. weight decay  &      $1e-8$   &      $1e-8$       \\
Param. warm up       &   False   &   False         \\
Alpha learning rate  &   $0.001$ &   $0.0001$            \\
Alpha weight decay   &   $0.001$    &   $0.0001$             \\
Alpha warm up        &     True   &     True       \\
Alpha scheduler      &       Linear     &       Linear      \\
Alpha optimizer      &     Gradient Descent  &     Gradient Descent       \\
\bottomrule
\end{tabular}
}

\end{table}

\begin{table*}
\centering
\caption{BOHB optimized hyperparameters for different data formations, objectives and methods. \label{table:HP_BOHB}}
\tiny
\resizebox{\textwidth}{!}{
\begin{tabular}{@{}lccccc@{}}
\toprule
Hyperparameter       & Search Range &  BOHB-one-shot-Blur & BOHB-one-shot-DS & BOHB-Blur & BOHB-DAS-single\\ \midrule
Param. learn. rate & $[1e-05,1]$ & $0.0014232405$   & $0.0020448382$       & $0.0020882283$  & $0.0014232405$  \\
Param. weight decay  & $[1e-08, 0.1]$  &   $8.616e-07$   &   $5.04e-08$    &   $4.4e-08$  & $8.616e-07$   \\
Param. warm up       &  [True,False] & False  & True  & False   & False   \\
Alpha learn. rate  &  $[1e-05, 0.1]$ & $0.0836808765$ & $0.0100063746$        & $8.43195e-05$  &  $0.025012337102395577$  \\
Alpha weight decay   &   $[1e-05, 0.1]$ & $5.05099e-05$ &$0.0058022776$     &$0.0127425783$     & $1.390640076980444e-05$      \\
Alpha warm up        &    [True, False ]& False    & True   & True  & False   \\
Alpha scheduler      &   [None, Linear] &   Linear &   Linear     &   Linear & None  \\
Alpha optimizer      &  [Adam, Gradient Descent]  & Adam  & Gradient Descent    & Adam   & Adam    \\
\bottomrule
\end{tabular}}

\end{table*}

\begin{table*}
\centering
\caption{BOHB optimized hyperparameters for the non-sequential search space for data formation blur and different objectives.\label{table:HP_BOHB_wide}}
\resizebox{\textwidth}{!}{
\begin{tabular}{@{}lccc@{}}
\toprule
Hyperparameter       & Search Range &   BOHB-Non-Seq-one-shot-Blur & BOHB-Non-Seq-Blur\\ \midrule
Param. learn. rate & $[1e-05,1]$  &$0.0050969066$ & $0.0037014752$\\
Param. weight decay  & $[1e-08, 0.1]$  &   $ 2.423e-07$ & $1.4573e-06$ \\
Param. warm up       &  [True,False] & False   &  False\\
Alpha learnn rate  &  $[1e-05, 0.1]$ &  $ 1.32499e-05$ & $0.0012395056$ \\
Alpha weight decay   &   $[1e-05, 0.1]$ &  $0.0010171142$&  $0.0002855732$  \\
Alpha warm up        &    [True, False ]  & False & False \\
Alpha scheduler      &   [None, Linear] &   None &   None\\
Alpha optimizer      &  [Adam, Gradient Descent]   & Adam    & Adam   \\
\bottomrule
\end{tabular}}
\end{table*}

\section{Computational Setup}
All experiments in the main body were run on a single \texttt{Nvidia GTX 2080ti} graphics card of which two were utilized. The hyperparameter tuning with BOHB was conducted on a single \texttt{Nvidia GTX 1080 Ti} graphics card.

\end{document}